\documentclass[11pt,letterpaper,twocolumn]{article}

\date{}

\usepackage[utf8]{inputenc}
\usepackage[T1]{fontenc}
\usepackage{hyperref}
\usepackage{url}
\usepackage{booktabs}
\usepackage{amsfonts}
\usepackage{nicefrac}
\usepackage{microtype}
\usepackage{graphicx}
\usepackage{natbib}
\usepackage{doi}
\usepackage{amsmath}
\usepackage{amssymb}
\usepackage{mathtools}
\usepackage{algorithm}
\usepackage{algpseudocode}
\usepackage{xcolor}
\usepackage{multirow}
\usepackage{tcolorbox}
\usepackage{float}

\usepackage[top=1in, bottom=1in, left=1in, right=1in]{geometry}

\definecolor{lightorange}{RGB}{255,236,217}
\definecolor{lightblue}{RGB}{217,236,255}

\title{An Empirical Comparison of Text Summarization: A Multi-Dimensional Evaluation of Large Language Models}

\author{
  Anantharaman Janakiraman, Behnaz Ghoraani, Ph.D. \\
  \textit{Department of Electrical Engineering and Computer Science,}  
  \textit{Florida Atlantic University} \\
  \texttt{ajanakiraman2024@fau.edu, bghoraani@fau.edu} \\
}

\begin{document}

\maketitle

\begin{abstract}
Text summarization is crucial for mitigating information overload across journalism, medicine, legal, and business intelligence. This research comprehensively evaluates text summarization performance across 17 large language models (OpenAI, Google, Anthropic, Open Source). We performed a novel multi-dimensional evaluation of these models using seven diverse datasets (BigPatent, BillSum, CNN/DailyMail, PubMed, SAMSum, WikiHow, XSum) at three output lengths (50, 100, 150 tokens) and several evaluation metrics including factual consistency, semantic similarity, lexical overlap, and human-like evaluation. Our study also considered both quality and efficiency factors to assess real-world deployment feasibility. The key findings from our extensive analysis reveal significant differences between models, with specific models excelling in factual accuracy (deepseek-v3), human-like quality (claude-3-5-sonnet), processing efficiency (gemini-1.5-flash or gemini-2.0-flash), and cost effectiveness (gemini-1.5-flash or gemini-2.0-flash). Performance varies dramatically by dataset, with models struggling on technical domains (BigPatent, PubMed) but performing well on conversational content (SAMSum). We identified optimal summary lengths for different metrics, revealing a critical tension between factual consistency (best at 50 tokens) and perceived quality (best at 150 tokens). Our analysis provides evidence-based recommendations for different use cases, from high-stakes applications requiring factual accuracy to resource-constrained environments needing efficient processing. Our findings establish a crucial foundation for advancing next-generation summarization systems. This comprehensive analysis significantly enhances evaluation methodology by integrating both quality metrics and operational considerations, incorporating critical trade-offs between accuracy, efficiency, and cost-effectiveness that should provide guidance for evidence-based model selection tailored to specific application requirements.
\end{abstract}


\section{Introduction}
The exponential growth of digital text has made automated text summarization increasingly vital across domains: journalism, academia, business intelligence, healthcare, and so on. Text summarization systems condense information in source documents while preserving key information, enabling the user to understand essential content without the need to read all source documents. Although recent progress in large language models (LLMs) have shown significant improvement in their summarization ability, there is still a great deal to be done in developing comprehensive evaluation frameworks, which account for multiple quality dimensions in conjunction with practical concerns of deployment.

Traditional evaluation methods in text summarization have relied heavily on lexical overlap metrics (e.g., ROUGE \cite{reference13}), which measure word-level similarity between generated summaries and human references. Although these metrics provide valuable information, they do not capture semantic equivalence, factual consistency, and other critical dimensions of summary quality. Some recent work has proposed metrics such as BERTScore \cite{reference11} for semantic similarity, and SummaC \cite{reference12} for factual consistency, but most comparative analyses focus on a single dimension of quality, or do not consider practical considerations for deployment, such as processing efficiency and cost.

This research addresses these limitations by proposing a balanced multidimensional evaluation framework that assigns appropriate weights to factual consistency (35\%), semantic similarity (25\%), lexical overlap (15\%), and human-like evaluation (25\%). In addition, we incorporate a 70/30 quality-efficiency split to assess practical deployment considerations. Our comprehensive evaluation spans 17 large language models, including variants from OpenAI, Anthropic, Google, and open-source models, across seven diverse datasets and three different output lengths. 

Our findings reveal important performance variations across models, with some LLMs exhibiting superior capabilities in specific dimensions such as factual accuracy, human-like quality, processing efficiency, and cost-effectiveness. Additionally, we observe that the optimal summary length varies based on the prioritized evaluation metric and the specific application context. This research makes several contributions to the text summarization evaluation by: 
\begin{itemize}
    \item Conducting a comprehensive comparative analysis of 17 state-of-the-art LLMs using a balanced multidimensional framework.
    \item Offering evidence-based insights into the relationship between quality metrics and practical deployment considerations.
    \item Providing detailed recommendations for model selection across different use cases
    \item Establishing a replicable evaluation methodology that better aligns with real-world requirements
\end{itemize}
These contributions delineate the innovative aspects of our work and its practical implications, underscoring how our research extends the boundaries of text summarization evaluation. The remainder of this paper is structured as follows. Section 2 presents related works on text summarization and evaluation methods; Section 3 discusses our evaluation system and setup; Section 4 shows results of the comparison; Section 5 highlights implications of our analysis and provides recommendations; and Section 6 ends with limitations and future research directions.

\section{Background and Related Work}
The task of text summarization has evolved significantly across the last few decades, from its early extractive methods that consist of selecting the important sentences verbatim from the source to its recent abstractive approaches that generate novel text while retaining key information from the document. The field has seen dramatic advancements with the rise of large language models (LLMs), which have demonstrated remarkable capabilities in generating coherent and contextually relevant summaries.

\subsection{Summarization Methods}
Summarization approaches span several key dimensions that define their functioning and output characteristics. Extractive methods identify and select important sentences verbatim from source documents, while abstractive approaches generate new text that conveys essential information through reformulation. Summarization can target either single documents, focusing on one source text, or multiple documents, requiring content selection across sources and resolution of potentially conflicting information. The output intent varies between generic summarization, which creates general overviews of document content, and query-focused summarization, which addresses specific user information needs. Additionally, summaries may be indicative, providing a brief overview of what the document contains, or informative, generating comprehensive content that covers all key points in the source material.

\subsection{Evolution of Summarization}
Traditional summarization systems relied on statistical features and rule-based methods to extract important sentences from source documents. These extractive methods, while effective for basic summarization, often produced disjointed outputs lacking coherence and contextual flow \cite{reference1}. The advent of neural networks, particularly sequence-to-sequence models with attention mechanisms, marked a significant shift toward abstractive summarization capabilities.

With the advent of transformer architectures and pre-trained language models, there has been a fundamental shift in generating summaries with improved coherence, grammatical correctness, and content selection. However, these advancements brought a new set of challenges, particularly around factual consistency, the tendency of models to generate content that contradicts or misrepresents the source material \cite{reference12}.

Evaluating summarization systems presents inherent challenges due to the subjectivity in defining "good" summaries. Several evaluation approaches have emerged: \textit{Automated Reference-based Metrics} compare generated summaries against human-written references, with traditional metrics like ROUGE (Lin, 2004) emphasizing lexical overlap and newer methods like BERTScore (Zhang et al., 2019) capturing semantic similarity; \textit{Reference-Free Evaluation} assesses summaries without human references, including factual consistency measures like SummaC (Laban et al., 2022c) that detect contradictions; \textit{LLM-Based Evaluation} leverages models like GPT-4 or Claude to assess dimensions such as coherence and factuality, offering a scalable middle ground between automated metrics and human judgment; \textit{Human Evaluation} remains the gold standard despite being resource-intensive, with evaluators assessing relevance, coherence, and readability; and \textit{Efficiency and Deployment Metrics} consider practical factors like processing time, computational requirements, and operational costs, which are increasingly critical for real-world applications.

Most prior comparative studies have focused only on certain quality metrics (often ROUGE scores) without fully considering a broader evaluation landscape or deployment aspects. Our study aims to bridge this gap by introducing a balanced framework that incorporates multiple quality dimensions alongside efficiency metrics, enabling an assessment that is more representative of practical real-world requirements.

\subsection{Comparative Studies and Balanced Evaluation}
Several prior studies have compared summarization capabilities across different models. However, these comparisons focus mainly on one dimension of quality (e.g., ROUGE scores or BERT scores) without considering the multifaceted nature of summarization quality. The challenge of balancing different quality aspects with practical deployment considerations remains largely unaddressed in the literature.

Our study builds on previous work while providing a more comprehensive evaluation framework. In this study, we assign appropriate weights to multiple quality dimensions (factual consistency, semantic similarity, lexical overlap, and human-like quality) alongside efficiency metrics and our aim is to perform a comprehensive assessment of summarization capabilities across diverse LLMs. This approach aligns better with real-world requirements, where model selection must consider both quality and practical constraints.

\section{Methodology}

Our research employs a systematic approach to evaluate text summarization capabilities across a diverse range of models, datasets, and output lengths. This section details our experimental setup, including model selection, datasets, evaluation metrics, and procedural details.

\subsection{Models}

We evaluate 17 models representing a diverse range of architectures, capabilities, and accessibility. Table \ref{tab:models} provides an overview of the models included in our evaluation.

\begin{table*}[h!]
\centering
\small
\begin{tabular}{@{}llll@{}}
\toprule
\textbf{Model Family} & \textbf{Model Name} & \textbf{Type} & \textbf{Access Method} \\
\midrule
\multirow{3}{*}{Anthropic} & claude-3-5-haiku & Commercial & API \\
 & claude-3-5-sonnet & Commercial & API \\
 & claude-3-opus & Commercial & API \\
\midrule
\multirow{3}{*}{Google} & gemini-1.5-flash & Commercial & API \\
 & gemini-1.5-pro & Commercial & API \\
 & gemini-2.0-flash & Commercial & API \\
\midrule
\multirow{1}{*}{DeepSeek} & deepseek-v3 & Commercial & API \\
\midrule
\multirow{4}{*}{OpenAI} & gpt-3.5-turbo & Commercial & API \\
 & gpt-4-turbo & Commercial & API \\
 & gpt-4o & Commercial & API \\
 & gpt-4o-mini & Commercial & API \\
 & o1 & Commercial & API \\
 & o1-mini & Commercial & API \\ 
\midrule
\multirow{7}{*}{Open-source} & deepseek-7b & Open-source & Local inference \\
 & falcon-7b & Open-source & Local inference \\
 & llama-3.2-3b & Open-source & Local inference \\
 & mistral-7b & Open-source & Local inference \\
\bottomrule
\end{tabular}
\caption{Overview of evaluated models and their access methods}
\label{tab:models}
\end{table*}

This selection encompasses a range of model sizes, architectures, and training approaches, allowing us to identify performance patterns across different model families and scales.

\subsection{Multi-Dimensional Evaluation Framework}

\begin{figure*}[t]
\centering
\includegraphics[width=0.7\textwidth]{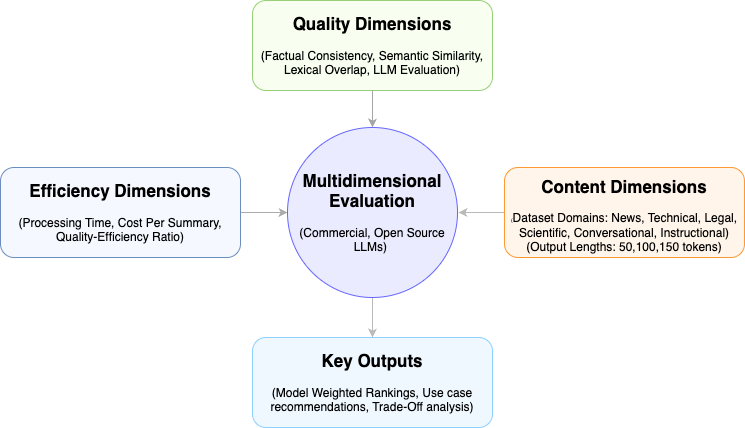}
\caption{Multi-dimensional evaluation framework for assessing large language models on text summarization}
\label{fig:multidimension-eval}
\end{figure*}

Text summarization quality cannot be adequately assessed through a single metric or under a single condition. Our research employs a comprehensive multi-dimensional evaluation framework (Figure~\ref{fig:multidimension-eval}) that systematically evaluates LLM performance across three key dimensions: quality, efficiency, and content. By analyzing these dimensions simultaneously, we can identify complex trade-offs and interactions that would not be apparent from single-dimension evaluations, such as how factual consistency varies with summary length across different domains, or how efficiency considerations influence model selection for specific use cases.

\subsubsection{Datasets}

To ensure comprehensive evaluation across diverse domains and summarization challenges, we selected seven datasets representing different text types, styles, and complexity levels. Table~\ref{tab:datasets} provides an overview of these datasets, which span news (CNN/Daily Mail, XSum), technical documentation (BigPatent), legal texts (BillSum), scientific literature (PubMed), conversational dialogues (SAMSum), and instructional content (WikiHow).

For each dataset, we randomly sampled 30 documents to ensure feasible evaluation while maintaining representative coverage across document lengths and complexity levels. For particularly long source documents (BigPatent and PubMed), we truncated inputs to 4,096 tokens to accommodate model context window limitations while preserving essential content.

These diverse datasets enable us to evaluate multiple performance dimensions: domain adaptation (handling specialized terminology), text length handling (from brief dialogues to lengthy technical documents), abstractive capacity (generating novel phrasings rather than verbatim extraction), and information density (effectiveness across varying compression requirements).

\begin{table*}[h!]
\centering
\small
\begin{tabular}{@{}p{2cm}p{3cm}p{3cm}p{4cm}@{}}
\toprule
\textbf{Dataset} & \textbf{Domain} & \textbf{Characteristics} & \textbf{Summary Style} \\
\midrule
CNN/Daily Mail & News (journalism) & Multi-paragraph news articles covering a wide range of topics & Multi-sentence abstractive summaries capturing key points \\
\addlinespace
XSum & News (BBC) & News articles spanning various topics with diverse writing styles & Single-sentence, highly abstractive summaries requiring significant content compression \\
\addlinespace
BigPatent & Technical (patent documentation) & Technical documents with specialized terminology and complex structures & Technical abstracts with focus on novelty and application \\
\addlinespace
BillSum & Legal (U.S. congressional bills) & Legal and legislative language with complex structure and domain-specific terminology & Condensed summaries of legal provisions and implications \\
\addlinespace
PubMed & Scientific (biomedical literature) & Scientific articles with specialized medical terminology and research methodology & Structured scientific abstracts following academic conventions \\
\addlinespace
SAMSum & Conversational (dialogues) & Messenger-like conversations between multiple participants with informal language & Third-person narrative summarizing key points of conversation \\
\addlinespace
WikiHow & Instructional (how-to guides) & Step-by-step procedures across various topics with instructional intent & Concise summaries of procedural knowledge highlighting core steps \\
\bottomrule
\end{tabular}
\caption{Detailed characteristics of datasets used in the evaluation}
\label{tab:datasets}
\end{table*}

\subsubsection{Evaluation Metrics}

Our evaluation framework employs a balanced multi-dimensional approach that captures different aspects of summarization quality while also considering practical deployment factors. We carefully calibrated weights across several metric categories to ensure a comprehensive assessment that aligns with real-world requirements for summarization systems.

\paragraph{Quality Metrics}

Text summarization quality is inherently multifaceted, requiring measurement across several complementary dimensions. Traditional metrics like ROUGE capture lexical overlap but fail to account for semantic equivalence or factual accuracy. To address these limitations, our evaluation framework incorporates multiple metric families with carefully balanced weights that reflect their relative importance in practical applications.

Our weighting scheme prioritizes factual consistency (35\%) as the most critical aspect of summarization quality, particularly for high-stakes applications where misinformation could have significant consequences. Semantic similarity (25\%) captures meaning preservation beyond exact word matching, while traditional lexical overlap metrics (15\%) provide continuity with established benchmarks. Finally, our LLM-based evaluation (25\%) introduces a human-like assessment of subjective aspects like coherence and readability that automated metrics often miss. These weights are summarized in Table~\ref{tab:quality-metrics}.

\begin{table*}[h!]
\centering
\begin{tabular}{@{}llc@{}}
\toprule
\textbf{Metric Category} & \textbf{Representative Measures} & \textbf{Weight} \\
\midrule
Factual Consistency & SummaC scores & 35\% \\
Semantic Similarity & BERTScore F1 & 25\% \\
Lexical Overlap & ROUGE-1, ROUGE-2, ROUGE-L & 15\% \\
Human-like Quality & LLM-based evaluation & 25\% \\
\bottomrule
\end{tabular}
\caption{Quality evaluation metrics with balanced weights}
\label{tab:quality-metrics}
\end{table*}

\paragraph{Automated Metrics Details}

Our evaluation employs complementary metrics to capture different quality aspects. For lexical overlap, we use ROUGE \cite{reference13} to measure n-gram similarity via ROUGE-1 (unigram overlap for content coverage), ROUGE-2 (bigram overlap for fluency), and ROUGE-L (longest common subsequence for structural similarity), computing precision, recall, and F1 scores with F1 as our primary metric despite its limitation in capturing semantic equivalence. For semantic similarity, BERTScore \cite{reference11} addresses these limitations by representing tokens with BERT embeddings, computing cosine similarity between semantically equivalent but lexically different tokens, applying content-word weighting, and reporting precision, recall, and F1 scores, with F1 as our primary semantic metric due to its stronger correlation with human judgments for abstractive summarization. For factual consistency, SummaC \cite{reference12} employs Natural Language Inference models to detect source-summary contradictions, using a convolutional layer (SummaC-Conv) to aggregate token-level NLI signals into document-level consistency scores between 0-1 (higher indicating greater factual alignment), operating reference-free against source documents to detect hallucinations missed by similarity metrics.

\paragraph{LLM-based Evaluation}

To complement automated metrics with human-like assessment, we employed LLM-based evaluation using either GPT-3.5-Turbo or Claude-3.5-Haiku as evaluators for 10 samples per model-dataset-length configuration, ensuring the evaluator differed from the model being evaluated to prevent self-evaluation bias. We used carefully crafted prompts providing clear evaluation instructions with explicit criteria, the full source text (truncated to 4,000 characters if necessary), the generated summary, and a structured output format requirement. Each summary was assessed on five dimensions (relevance, coherence, factual consistency, conciseness, and overall quality) using a 1-5 scale, with evaluators returning ratings in a structured JSON format. To ensure evaluation consistency, we used temperature=0.1 to minimize randomness, implemented error handling for malformed outputs, and performed quality checks on evaluation subsets to verify alignment with human judgments. This approach provided crucial insights into subjective quality dimensions that automated metrics often miss.

\paragraph{Efficiency Metrics}

In addition to quality metrics, we gathered efficiency data to address practical deployment considerations. For processing time, we measured end-to-end performance for each summary generation, including API request-to-response time (incorporating network latency) for API-based models and full inference time for open-source models, reporting average time in seconds per summary with standard deviation. Time measurements used consistent hardware configurations (NVIDIA A100 GPUs) for open-source models and similar time periods for API-based models to minimize external variance. 

For cost estimation, we calculated direct costs for API-based models using precise input/output token counts, official pricing tiers as of February 2025, and separate tracking for prompt and completion costs to reflect different pricing structures. For open-source models, we estimated operational costs based on GPU-hours (approximately \$0.8/hour for A100), normalized by throughput for per-summary costs, with added overhead factors for system maintenance. 

To incorporate efficiency into our ranking system, we inverted processing time and cost metrics (1 - normalized value) so lower values received higher scores, normalized all values to a 0-1 range for consistent weighting with quality metrics, and calculated combined efficiency scores as weighted averages of normalized time and cost, providing crucial context for resource-constrained or high-volume applications.

\subsection{Experimental Setup}

For each model-dataset pair, we generated summaries at three different output lengths (50, 100, and 150 tokens) to analyze length effects on quality and efficiency. Our procedure involved processing 30 examples from each dataset per model and token length combination, using consistent minimal prompts across all models for fair comparison, with temperature setting of 0.1 for deterministic outputs. API-based models used their respective service providers, while open-source models were evaluated on NVIDIA A100 GPUs. For each summary, we computed all automated metrics, collected timing information, and tracked token usage to calculate costs using published pricing for API-based models or estimated computational costs based on GPU time for open-source models.

\subsection{Ranking Methodology}

\begin{figure}
    \centering
    \includegraphics[width=0.7\linewidth]{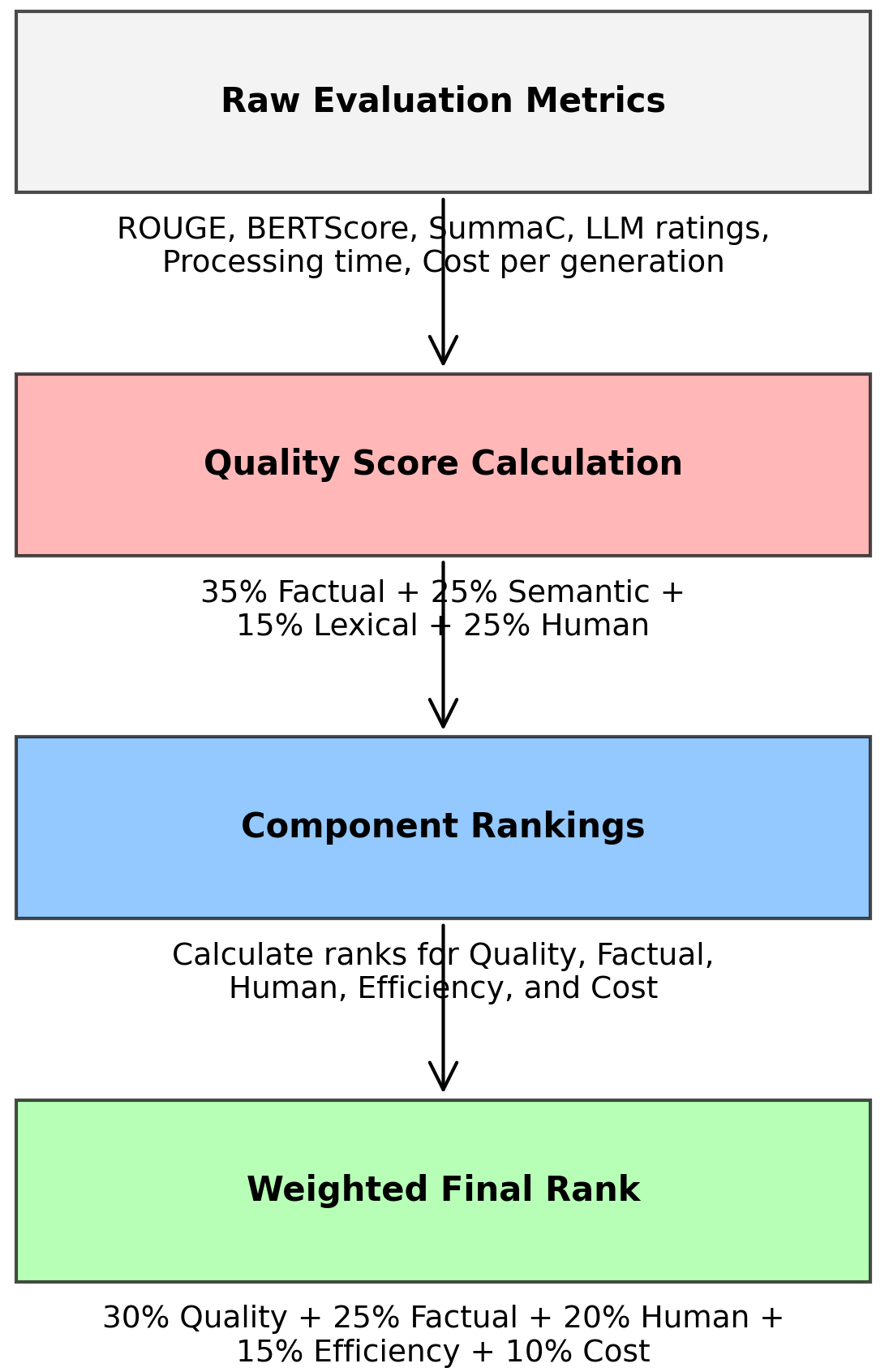}
    \caption{Ranking Process}
    \label{fig:ranking-process}
\end{figure}

\begin{table*}[h!]
\centering
\begin{tabular}{@{}lcc@{}}
\toprule
\textbf{Ranking Component} & \textbf{Weight} & \textbf{Description} \\
\midrule
Quality Rank & 30\% & Combined quality metrics rank \\
Factual Consistency Rank & 25\% & Specific emphasis on factual consistency \\
Human-like Evaluation Rank & 20\% & LLM-based evaluation importance \\
Efficiency Rank & 15\% & Processing time considerations \\
Cost Efficiency Rank & 10\% & Budget impact for production systems \\
\bottomrule
\end{tabular}
\caption{Ranking components with balanced weights}
\label{tab:ranking-weights}
\end{table*}

To produce comprehensive rankings balancing quality and efficiency, we implemented a two-level weighting scheme as shown in Figure~\ref{fig:ranking-process} and Table~\ref{tab:ranking-weights}. First, we ranked models separately on different components (quality metrics, factual consistency, human-like evaluation, efficiency, and cost). Second, we combined these component rankings into a final score using our balanced weights: quality rank (30\%), factual consistency rank (25\%), human-like evaluation rank (20\%), efficiency rank (15\%), and cost efficiency rank (10\%). Additionally, for applications requiring both quality and efficiency, we applied a quality-efficiency trade-off analysis with a 70/30 split (70\% weighted toward combined quality metrics and 30\% toward combined efficiency metrics). This balanced approach identifies models excelling in specific dimensions while recognizing those providing the best overall value across multiple evaluation criteria, reflecting the practical reality that most applications prioritize output quality while considering computational and financial constraints.

\subsection{Implementation Details}

The evaluation pipeline was implemented in Python, utilizing HuggingFace's \texttt{transformers} and \texttt{datasets} libraries for model access and data loading, SummaC implementation for factual consistency evaluation, BERTScore and ROUGE implementations for semantic and lexical evaluation, API clients for commercial models (OpenAI, Anthropic, Google, DeepSeek), and PyTorch for local model inference. For reproducibility, we have open-sourced our evaluation code and detailed configuration settings, with the complete implementation allowing for extension to additional models, datasets, and metrics as they become available.

\begin{figure*}[t!]
\centering
\includegraphics[width=0.95\textwidth]{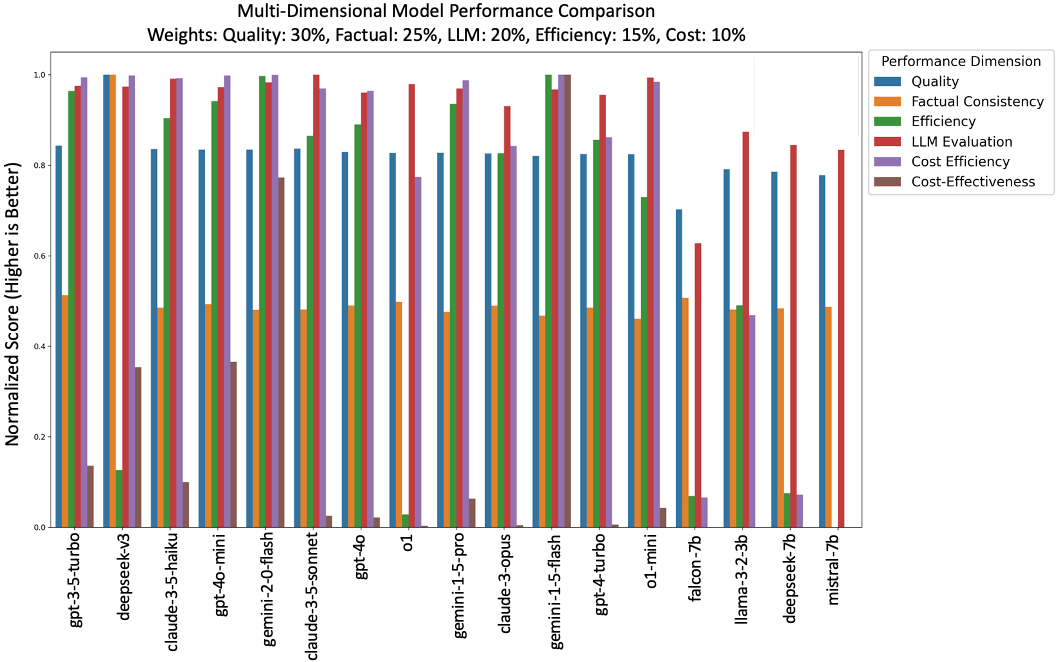}
\caption{Comparative analysis of 17 models across all evaluation dimensions, normalized to the 0-1 range for fair comparison.}
\label{fig:overall-performance}
\end{figure*}

\section{Results}

\subsection{Overall Performance}

\begin{table*}[t]
\centering
\small
\begin{tabular}{@{}lccccc@{}}
\toprule
\textbf{Model} & \textbf{Quality} & \textbf{Factual} & \textbf{Efficiency} & \textbf{Cost-Effectiveness} & \textbf{Weighted Rank} \\
\midrule
gpt-3.5-turbo & 2 & 2 & 3 & 5 & 3.25 \\
deepseek-v3 & 1 & 1 & 13 & 4 & 4.30 \\
claude-3-5-haiku & 4 & 9 & 6 & 6 & 5.55 \\
gpt-4o-mini & 6 & 5 & 4 & 3 & 5.55 \\
gemini-2.0-flash & 5 & 14 & 2 & 2 & 6.30 \\
\bottomrule
\end{tabular}
\caption{Top 5 models by weighted rank across all metrics. Lower ranks are better (1 = best). The weighted rank combines the component ranks according to our consistent weighting system: Quality (30\%), Factual (25\%), Efficiency (15\%), and Cost-Effectiveness (10\%).}
\label{tab:overall}
\end{table*}

Based on our balanced evaluation framework that prioritizes quality (30\%), factual consistency (25\%), human-like evaluation (20\%), efficiency (15\%), and cost (10\%), we find significant differences in performance across the 17 evaluated models. As shown in Table~\ref{tab:overall}, gpt-3.5-turbo emerges as the top overall performer with a weighted rank of 3.25, excelling in both quality metrics and factual consistency while maintaining good efficiency. Notably, deepseek-v3 achieves the second position despite ranking lower on efficiency, primarily due to its exceptional performance in factual consistency. Figure~\ref{fig:overall-performance} shows the performance of all evaluated models

\subsection{Factual Consistency}

\begin{table}[t]
\centering
\small
\begin{tabular}{@{}lc@{}}
\toprule
\textbf{Model} & \textbf{SummaC Score} \\
\midrule
deepseek-v3 & 0.6823 \\
gpt-3.5-turbo & 0.3501 \\
falcon-7b & 0.3460 \\
o1 & 0.3399 \\
gpt-4o-mini & 0.3363 \\
\bottomrule
\end{tabular}
\caption{Models with highest factual consistency}
\label{tab:factual}
\end{table}

Factual consistency, weighted at 35\% in our quality metrics framework, shows particularly interesting patterns. As shown in Table~\ref{tab:factual}, deepseek-v3 dramatically outperforms all other models with a SummaC score of 0.6823, which is 94.9\% higher than the next best model (gpt-3.5-turbo at 0.3501). This exceptional performance in factual consistency suggests that deepseek-v3 has been specifically optimized to avoid generating content that contradicts or misrepresents the source material.

The remaining top models for factual consistency show relatively similar scores, clustering in the 0.33-0.35 range, with both commercial API models (gpt-3.5-turbo, gpt-4o-mini, o1) and open-source models (falcon-7b) represented. This suggests that both proprietary and open-source approaches can achieve comparable levels of factual reliability.

\subsection{Human-like Evaluation}

\begin{table}[t]
\centering
\small
\begin{tabular}{@{}lc@{}}
\toprule
\textbf{Model} & \textbf{LLM Score} \\
\midrule
claude-3-5-sonnet & 4.75 \\
o1-mini & 4.72 \\
claude-3-5-haiku & 4.70 \\
gemini-2.0-flash & 4.66 \\
o1 & 4.65 \\
\bottomrule
\end{tabular}
\caption{Models with highest LLM-based evaluation scores (scale: 1-5)}
\label{tab:llm-eval}
\end{table}

For human-like quality assessment, measured through our LLM-based evaluation on a 1-5 scale and weighted at 25\% in our quality metrics, Anthropic's Claude models demonstrate superior performance. As shown in Table~\ref{tab:llm-eval}, claude-3-5-sonnet achieves the highest score (4.75/5.0), followed closely by o1-mini and claude-3-5-haiku. These results indicate that these models excel at generating summaries that human evaluators would likely judge as high-quality in terms of relevance, coherence, factual accuracy, and conciseness.

\subsection{Effect of Output Length}

\begin{table}[t]
\centering
\footnotesize
\begin{tabular}{@{}ccccc@{}}
\toprule
\textbf{Tokens} & \textbf{ROUGE-1} & \textbf{BERT} & \textbf{SummaC} & \textbf{LLM} \\
\midrule
50 & 0.241 & 0.857 & 0.486 & 4.28 \\
100 & 0.250 & 0.856 & 0.290 & 4.47 \\
150 & 0.249 & 0.854 & 0.281 & 4.51 \\
\bottomrule
\end{tabular}
\caption{Average quality metrics by token length}
\label{tab:length}
\end{table}

\begin{figure*}[h!]
    \centering
    \includegraphics[width=0.95\textwidth]{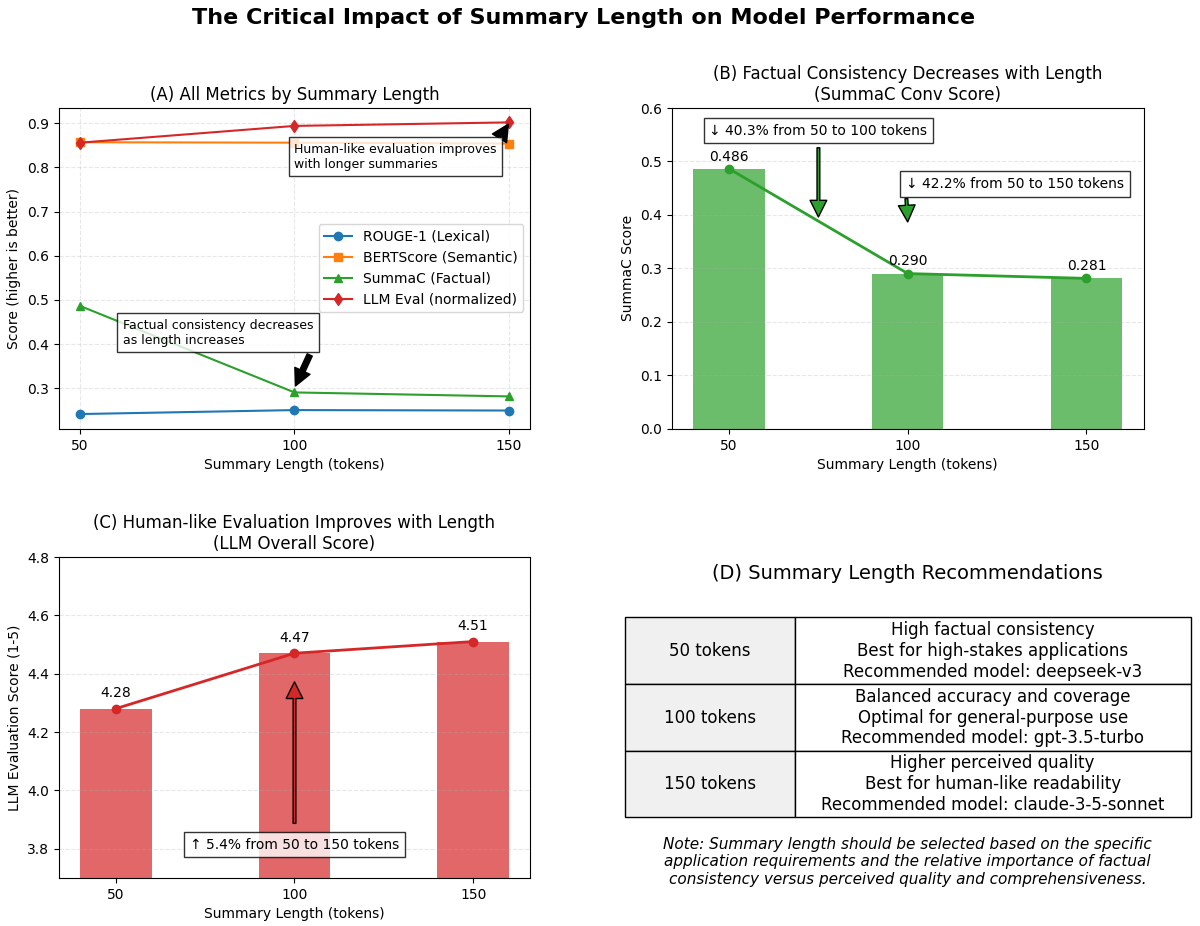}
    \caption{Impact of Summary Length on Performance Metrics: This comprehensive visualization shows how different quality metrics vary with summary length (50, 100, and 150 tokens).}
    \label{fig:length-impact}
\end{figure*}

Our analysis of output length effects, summarized in Table~\ref{tab:length}, reveals important trade-offs across different quality dimensions. We find that the optimal token length varies by metric, with ROUGE-1 F1 peaking at 100 tokens (0.250), suggesting this length optimizes content coverage, while BERTScore F1 slightly decreases as length increases, with best performance at 50 tokens (0.857). Most significantly, SummaC Conv is dramatically higher at 50 tokens (0.486) compared to longer summaries, indicating shorter summaries are significantly more factually consistent. In contrast, LLM Overall ratings increase with length, peaking at 150 tokens (4.51), suggesting human-like evaluators prefer more comprehensive summaries. This pattern reveals a critical tension between factual consistency and perceived quality: while human-like evaluators and lexical matching metrics like ROUGE prefer longer summaries, factual consistency is substantially better with shorter outputs. As shown in Figure~\ref{fig:length-impact}, ROUGE-1 F1 peaks at 100 tokens (0.250), suggesting this length optimizes content coverage. BERTScore F1 slightly decreases as length increases, with best performance at 50 tokens (0.857). Most significantly, SummaC Conv is dramatically higher at 50 tokens (0.486) compared to longer summaries, indicating shorter summaries are significantly more factually consistent. In contrast, LLM Overall ratings increase with length, peaking at 150 tokens (4.51), suggesting human evaluators prefer more comprehensive summaries.

\subsection{Performance by Dataset}

\begin{table*}[t]
\centering
\small
\begin{tabular}{@{}lcccc@{}}
\toprule
\textbf{Dataset} & \textbf{ROUGE-1 F1} & \textbf{BERTScore F1} & \textbf{SummaC Conv} & \textbf{LLM Overall} \\
\midrule
BigPatent & 0.288 & 0.853 & 0.093 & 4.34 \\
BillSum & 0.309 & 0.853 & 0.439 & 4.18 \\
CNN/DailyMail & 0.242 & 0.864 & 0.426 & 4.57 \\
PubMed & 0.225 & 0.833 & 0.102 & 3.93 \\
SAMSum & 0.307 & 0.883 & 0.531 & 4.77 \\
WikiHow & 0.195 & 0.846 & 0.397 & 4.62 \\
XSum & 0.160 & 0.857 & 0.480 & 4.55 \\
\bottomrule
\end{tabular}
\caption{Average quality metrics by dataset}
\label{tab:dataset-performance}
\end{table*}

Performance varies substantially across datasets (Table~\ref{tab:dataset-performance}), revealing important patterns in model capabilities. \textit{SAMSum} (conversational dialogues) shows the highest performance across most metrics, with top scores in BERTScore (0.883), SummaC (0.531), and LLM evaluation (4.77), suggesting models are particularly effective at summarizing dialogues. In contrast, \textit{BigPatent} and \textit{PubMed} yield notably lower factual consistency scores (0.093 and 0.102 respectively), indicating models struggle to maintain factual accuracy when summarizing technical content with specialized terminology. \textit{XSum} shows the lowest ROUGE-1 scores (0.160) but strong factual consistency (0.480), reflecting its highly abstractive single-sentence summary style that diverges lexically from references while maintaining factual alignment. \textit{BillSum} demonstrates strong lexical overlap (0.309 ROUGE-1) but lower human-like evaluation scores (4.18), suggesting that while models capture the content of legal documents, the summaries may lack coherence or readability. These domain-specific patterns highlight the importance of considering dataset characteristics when evaluating summarization performance and selecting models for specific applications.

\subsection{Efficiency Analysis}

\begin{table}[t]
\centering
\small
\begin{tabular}{@{}lcc@{}}
\toprule
\textbf{Model} & \textbf{Time (s)} & \textbf{Cost (\$)} \\
\midrule
gemini-1.5-flash & 1.08 & 0.00012 \\
gemini-2.0-flash & 1.14 & 0.00016 \\
gpt-3.5-turbo & 1.90 & 0.00108 \\
gpt-4o-mini & 2.41 & 0.00034 \\
gemini-1.5-pro & 2.55 & 0.00205 \\
\bottomrule
\end{tabular}
\caption{Top 5 models by processing time}
\label{tab:efficiency}
\end{table}

Processing efficiency, critical for real-world applications, shows significant variations across models (Table~\ref{tab:efficiency}). Google's Gemini models dominate the efficiency rankings, with gemini-1.5-flash leading at 1.08 seconds average processing time per summary. The efficiency advantage extends to cost metrics as well, with gemini-1.5-flash being the most cost-effective at approximately \$0.000124 per summary.

\begin{table}[t]
\centering
\small
\begin{tabular}{@{}lc@{}}
\toprule
\textbf{Model} & \textbf{Value Score} \\
\midrule
gemini-1.5-flash & 9696 \\
gemini-2.0-flash & 7493 \\
gpt-4o-mini & 3545 \\
deepseek-v3 & 3432 \\
gpt-3.5-turbo & 1322 \\
\bottomrule
\end{tabular}
\caption{Top 5 models by cost-effectiveness (Quality/Cost ratio)}
\label{tab:cost-effectiveness}
\end{table}

When examining cost-effectiveness (Table~\ref{tab:cost-effectiveness}), which considers the ratio of quality to cost, the Gemini models again demonstrate exceptional performance. Notably, deepseek-v3 appears in the top tier despite its lower processing efficiency, highlighting how its exceptional factual consistency provides value that offsets its higher computational requirements.

\subsection{Quality-Efficiency Trade-offs}

\begin{figure*}[t!]
    \centering
    \includegraphics[width=0.95\textwidth]{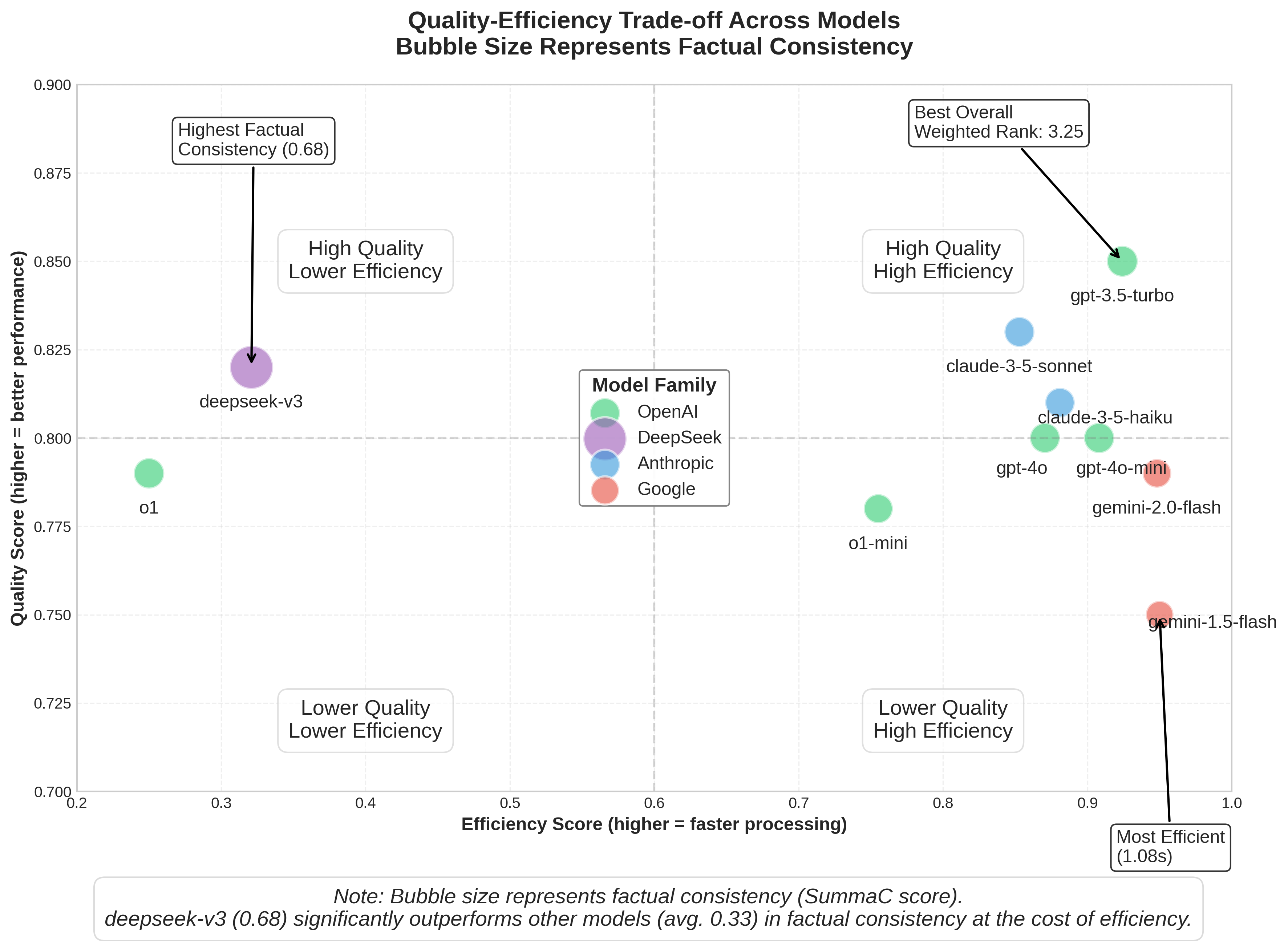}
    \caption{Quality-Efficiency Trade-off: Bubble size represents factual consistency score; position shows the balance between quality (y-axis) and efficiency (x-axis) metrics.}
    \label{fig:quality-efficiency}
\end{figure*}

\begin{table}[t]
\centering
\small
\begin{tabular}{@{}lc@{}}
\toprule
\textbf{Model} & \textbf{Balance Score} \\
\midrule
deepseek-v3 & 0.738 \\
gpt-3.5-turbo & 0.720 \\
gemini-2.0-flash & 0.718 \\
gpt-4o-mini & 0.714 \\
claude-3-5-haiku & 0.712 \\
\bottomrule
\end{tabular}
\caption{Models with best balance of quality (70\%) and efficiency (30\%)}
\label{tab:balance}
\end{table}

For real-world deployments, the balance between quality and efficiency is essential. Using a 70/30 quality-efficiency split (Table~\ref{tab:balance}), deepseek-v3 emerges as the model with the best overall balance (0.738), followed closely by gpt-3.5-turbo (0.720) and gemini-2.0-flash (0.718). This balanced perspective provides a more holistic view of model capabilities that better reflects practical deployment considerations.

This balanced perspective, illustrated in Figure~\ref{fig:quality-efficiency}, provides a more holistic view of model capabilities that better reflects practical deployment considerations. The visualization highlights the trade-offs between different dimensions, with some models prioritizing quality at the expense of efficiency (deepseek-v3) and others emphasizing speed with competitive quality (gemini models).

\section{Discussion}
\subsection{Model Specialization}
Our analysis reveals distinct patterns of model specialization across dimensions and domains. While most models perform best on the SAMSum dataset (dialogue summarization), deepseek-v3 shows particular strength on BigPatent, suggesting domain-specific optimization for technical content. This specialization indicates that models may be trained or optimized differently depending on their intended use cases.

The exceptional factual consistency of deepseek-v3 is particularly noteworthy. With a SummaC score of 0.6823, nearly double that of other models, it represents a significant advancement in faithful summarization. This performance comes at the cost of efficiency, however, as deepseek-v3 ranks 13th in processing time (21.03 seconds per summary). This trade-off exemplifies the tension between quality and efficiency that practitioners must navigate.

\subsection{Length-Quality Relationships}

Our findings regarding token length reveal critical trade-offs that should inform summary length selection. The dramatic decrease in factual consistency as length increases (from 0.486 at 50 tokens to 0.281 at 150 tokens) suggests that longer summaries increase the risk of hallucinations or factual errors. The opposing trend in human-like evaluation scores, which favor longer summaries (4.51 at 150 tokens vs. 4.28 at 50 tokens), highlights a potential disconnect between objective factual accuracy and subjective quality assessment. The peak in ROUGE scores at 100 tokens (0.250) suggests an optimal middle ground for content coverage. 

These patterns suggest that output length should be carefully calibrated based on the specific priorities of the application: shorter summaries for applications where factual accuracy is paramount, and longer summaries where perceived quality and comprehensiveness matter more. The inverse relationship between factual consistency and perceived quality represents a fundamental challenge in summarization system design, requiring practitioners to make deliberate choices about the appropriate length based on their specific use case requirements and quality priorities.

\subsection{Domain Challenges}

The substantial variation in performance across datasets highlights domain-specific challenges in text summarization. The poor factual consistency on technical domains (BigPatent: 0.093, PubMed: 0.102) compared to conversational content (SAMSum: 0.531) reveals a significant gap in models' ability to maintain factual accuracy when handling specialized terminology and complex concepts. This domain gap suggests a need for domain-specific fine-tuning or evaluation methods tailored to different content types. The pattern also suggests that practitioners should exercise particular caution when deploying general-purpose models for technical domains, where factual reliability is especially critical. These domain-specific performance disparities indicate that evaluating models on diverse datasets is essential for understanding their true capabilities and limitations, rather than relying on performance metrics from a single domain that may not generalize to other contexts. Future research should prioritize developing techniques to improve factual consistency specifically for technical and specialized domains where current models show the most significant weaknesses.

\subsection{Practical Recommendations}

Based on our comprehensive evaluation, we offer several use case-specific recommendations for text summarization model selection. For \textbf{high-stakes applications} where factual accuracy is critical, \texttt{deepseek-v3} provides unmatched factual consistency (SummaC: 0.6823). It is particularly suited for legal, medical, or financial summarization tasks where accuracy is paramount. To maximize factual reliability, output length should be limited to 50 tokens when possible. For applications prioritizing \textbf{human-like quality}, \texttt{claude-3-5-sonnet} achieved the highest human-like evaluation scores (4.75/5.0). It is recommended for contexts where subjective quality and readability are crucial, such as content intended for direct consumption by end users. For \textbf{general-purpose summarization}, \texttt{gpt-3.5-turbo} offers the best overall performance across balanced metrics (weighted rank: 3.25). This model is suitable for a broad range of summarization tasks, including content generation, document processing, and information retrieval. For \textbf{resource-constrained applications}, \texttt{gemini-1.5-flash} (1.08 seconds) or \texttt{gemini-2.0-flash} (1.14 seconds) deliver the fastest processing times and lowest costs ($0.000124 and $0.000163 per summary, respectively). These models are ideal for real-time applications, mobile platforms, or scenarios requiring high-volume processing. To achieve \textbf{optimal cost-effectiveness}, \texttt{gemini-1.5-flash} (score: 9695.99) followed by \texttt{gemini-2.0-flash} (score: 7493.17) provide the best balance between quality and cost. They are recommended for production environments with moderate quality demands and budget limitations. Lastly, for \textbf{domain-specific applications}, model selection should be guided by performance within the targeted domain. While most models perform effectively on conversational content (e.g., SAMSum), \texttt{deepseek-v3} demonstrates notable strengths in technical summarization tasks, particularly with specialized datasets like BigPatent.

\section{Limitations}
While our evaluation provides valuable insights, several factors should be considered when interpreting the results. Automated metrics, such as ROUGE scores, present inherent limitations in capturing valid yet lexically diverse summaries, and their reliance on human-written references may not reflect the full scope of valid summarization possibilities. Additionally, our sample size was relatively modest (30 examples per model-dataset-token length combination), suggesting that larger-scale evaluations might yield different trends or more robust findings. Human-like quality assessments, conducted using LLM-based evaluations with a limited subset (10 samples per configuration), serve as proxies and might not precisely mirror actual human judgments. Performance variations across different domains further indicate that our results may not generalize uniformly, underscoring the potential need for domain-adapted models. The weighting scheme adopted (35\% factual consistency, 25\% semantic similarity, 15\% lexical overlap, 25\% human-like evaluation) offers a balanced but potentially subjective representation of evaluation criteria. Finally, the rapid pace of model development implies that newer models released after this evaluation may exhibit improved capabilities, necessitating continual reassessment.

\section{Conclusion}
This study makes several significant contributions to the literature by introducing a comprehensive evaluation framework that integrates automated metrics and LLM-based human-like evaluations across multiple summarization models, datasets, and summary lengths. Our innovative approach provides a more nuanced and practically relevant assessment, addressing critical gaps in existing evaluation methodologies by emphasizing factual consistency, semantic similarity, readability, and efficiency. Notably, we identify key strengths of specific models—such as the superior factual reliability of deepseek-v3, the exceptional readability of claude-3-5-sonnet, and the robust versatility of gpt-3.5-turbo—offering actionable insights tailored to real-world applications. Additionally, the inclusion of processing time and cost-effectiveness considerations addresses practical deployment constraints. Future work should build upon our findings by enhancing factual consistency metrics, exploring advanced domain-adaptation techniques, evaluating newer models, and investigating hybrid human-model systems, further enriching the practical applicability and reliability of summarization technologies.

For additional detailed results and visualizations, please refer to the Appendix. The Appendix includes comprehensive ranking tables for all evaluated models, detailed quality metrics, model-specific visualizations across different evaluation dimensions, dataset-specific performance charts, and recommended configurations for various use cases.

\bibliographystyle{plainnat}
\nocite{*}
\bibliography{references}

\end{document}


\maketitle

\section{Full Ranking Table}

Table \ref{tab:full-rankings} provides the comprehensive ranking of all evaluated models across multiple dimensions, including quality, factual consistency, efficiency, and cost metrics.

\begin{table}[H]
\centering
\small
\setlength{\tabcolsep}{16pt} 
\begin{tabular}{@{}lccccc@{}}
\toprule
\textbf{Model} & \textbf{Quality} & \textbf{Factual} & \textbf{Efficiency} & \textbf{Cost-Effect.} & \textbf{Weighted} \\
& \textbf{Rank} & \textbf{Rank} & \textbf{Rank} & \textbf{Rank} & \textbf{Rank} \\
\midrule
gpt-3.5-turbo & 2 & 2 & 3 & 5 & 3.25 \\
deepseek-v3 & 1 & 1 & 13 & 4 & 4.30 \\
claude-3-5-haiku & 4 & 9 & 6 & 6 & 5.55 \\
gpt-4o-mini & 6 & 5 & 4 & 3 & 5.55 \\
gemini-2.0-flash & 5 & 14 & 2 & 2 & 6.30 \\
gemini-1.5-flash & 10 & 16 & 1 & 1 & 8.45 \\
gpt-4o & 3 & 7 & 7 & 14 & 8.80 \\
o1-mini & 8 & 6 & 11 & 9 & 9.00 \\
llama-3.2-3b & 12 & 10 & 12 & 7 & 10.65 \\
gpt-4-turbo & 7 & 11 & 9 & 15 & 11.30 \\
gemini-1.5-pro & 9 & 17 & 5 & 12 & 11.55 \\
claude-3-5-sonnet & 11 & 12 & 8 & 11 & 11.70 \\
o1 & 14 & 4 & 16 & 13 & 11.95 \\
claude-3-opus & 13 & 13 & 10 & 16 & 13.25 \\
falcon-7b & 15 & 3 & 15 & 17 & 13.40 \\
mistral-7b & 16 & 8 & 17 & 10 & 14.25 \\
deepseek-7b & 17 & 15 & 14 & 8 & 14.70 \\
\bottomrule
\end{tabular}
\caption{Full model rankings across quality dimensions, efficiency, and cost-effectiveness. Lower rank numbers indicate better performance (1 = best).}
\label{tab:full-rankings}
\end{table}

\section{Detailed Quality Metrics}

Table \ref{tab:all-quality} presents the raw scores for all quality metrics across models, providing a more granular view of performance.

\begin{table}[H]
\centering
\small
\begin{tabular}{@{}lcccccc@{}}
\toprule
\textbf{Model} & \textbf{ROUGE-1 F1} & \textbf{ROUGE-2 F1} & \textbf{BERTScore F1} & \textbf{SummaC} & \textbf{LLM Score} & \textbf{Time (s)} \\
\midrule
gpt-3.5-turbo & 0.251 & 0.089 & 0.860 & 0.350 & 4.58 & 1.90 \\
deepseek-v3 & 0.255 & 0.092 & 0.863 & 0.682 & 4.55 & 21.03 \\
claude-3-5-haiku & 0.249 & 0.088 & 0.858 & 0.318 & 4.70 & 3.27 \\
gpt-4o-mini & 0.247 & 0.087 & 0.856 & 0.336 & 4.60 & 2.41 \\
gemini-2.0-flash & 0.248 & 0.087 & 0.857 & 0.295 & 4.66 & 1.14 \\
gemini-1.5-flash & 0.242 & 0.083 & 0.853 & 0.289 & 4.54 & 1.08 \\
gpt-4o & 0.250 & 0.089 & 0.859 & 0.325 & 4.63 & 3.60 \\
o1-mini & 0.246 & 0.085 & 0.855 & 0.334 & 4.72 & 7.26 \\
llama-3.2-3b & 0.241 & 0.082 & 0.852 & 0.317 & 4.45 & 12.72 \\
gpt-4-turbo & 0.247 & 0.086 & 0.856 & 0.310 & 4.62 & 4.37 \\
gemini-1.5-pro & 0.245 & 0.084 & 0.854 & 0.284 & 4.57 & 2.55 \\
claude-3-5-sonnet & 0.241 & 0.083 & 0.852 & 0.307 & 4.75 & 4.16 \\
o1 & 0.240 & 0.081 & 0.851 & 0.340 & 4.65 & 23.28 \\
claude-3-opus & 0.241 & 0.082 & 0.852 & 0.303 & 4.60 & 5.04 \\
falcon-7b & 0.239 & 0.080 & 0.848 & 0.346 & 4.42 & 22.35 \\
mistral-7b & 0.238 & 0.079 & 0.847 & 0.322 & 4.36 & 23.93 \\
deepseek-7b & 0.237 & 0.078 & 0.846 & 0.292 & 4.32 & 22.20 \\
\bottomrule
\end{tabular}
\caption{Detailed quality metrics across all models}
\label{tab:all-quality}
\end{table}

\section{Model Weighted Rankings}

The weighted ranking combines all evaluation components according to our consistent weighting system to provide an overall assessment of model performance.

\begin{figure}[H]
    \centering
    \includegraphics[width=0.75\textwidth]{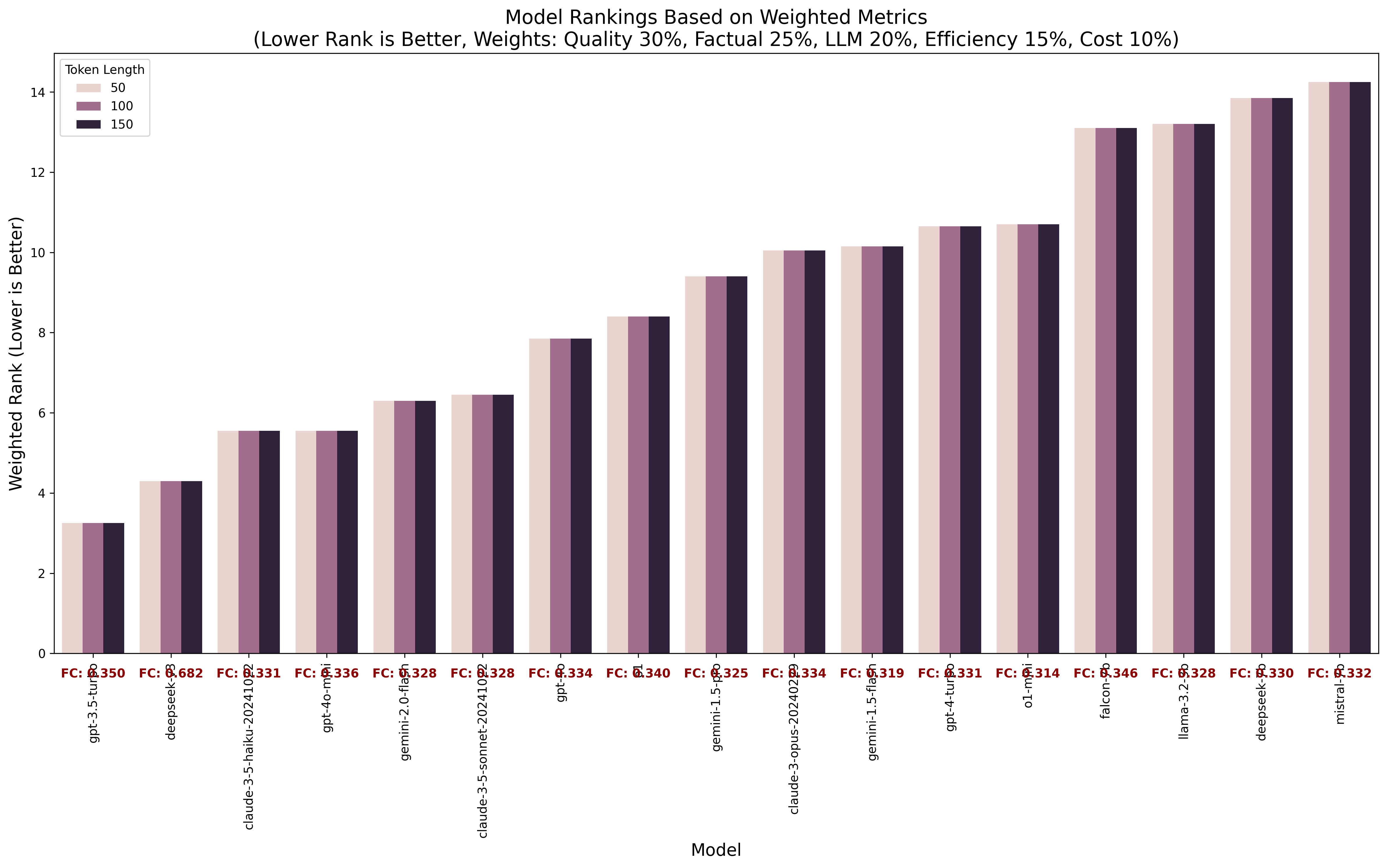}
    \caption{Models ranked by weighted score (lower is better)}
    \label{fig:weighted-rankings}
\end{figure}

\section{Factual Consistency Rankings}

\begin{figure}[H]
    \centering
    \includegraphics[width=0.75\textwidth]{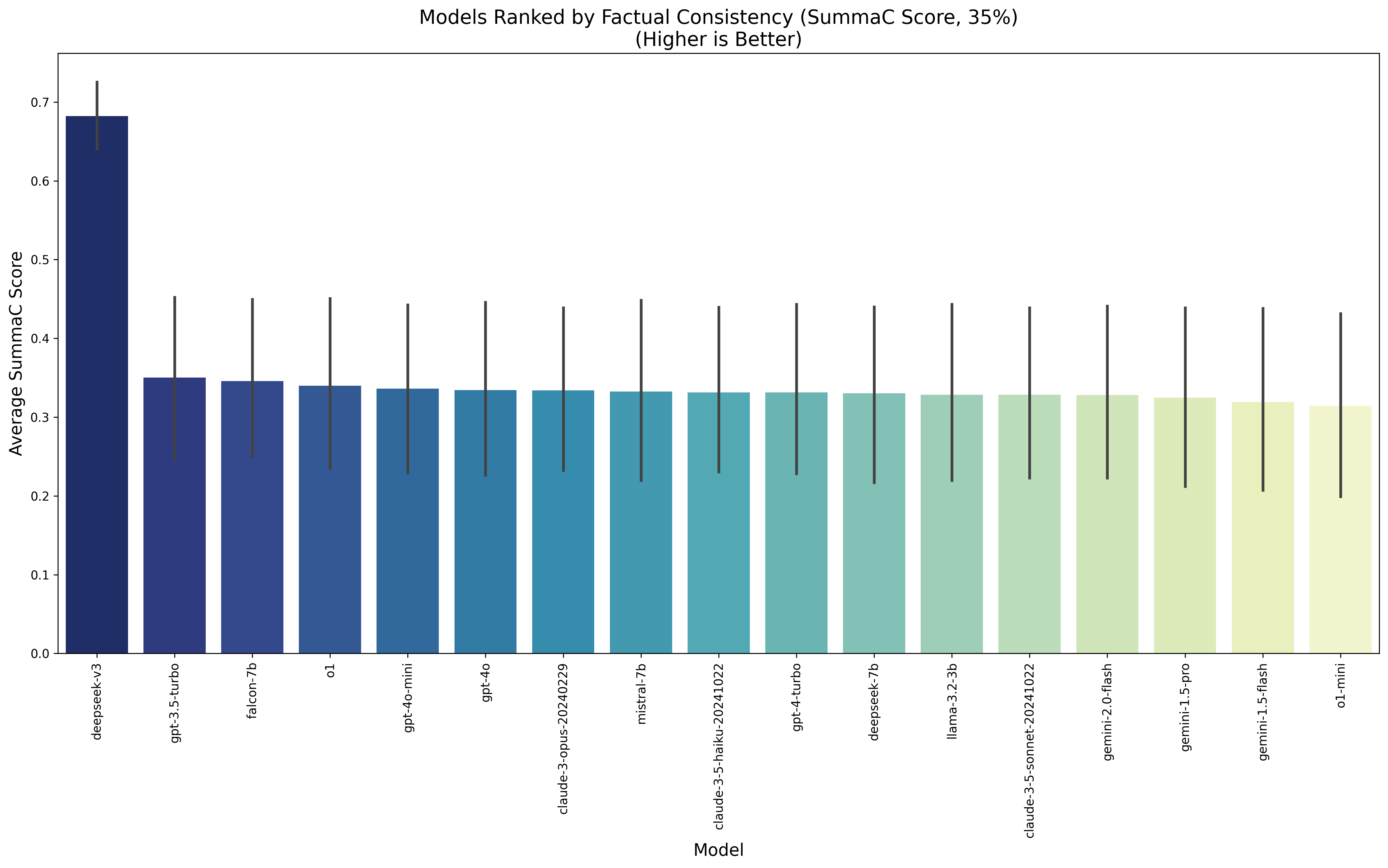}
    \caption{Models ranked by factual consistency}
    \label{fig:factual-ranking}
\end{figure}

Figure \ref{fig:factual-ranking} illustrates the ranking of models by factual consistency scores, highlighting the exceptional performance of deepseek-v3.

\section{LLM Evaluation Rankings}

\begin{figure}[H]
    \centering
    \includegraphics[width=0.75\textwidth]{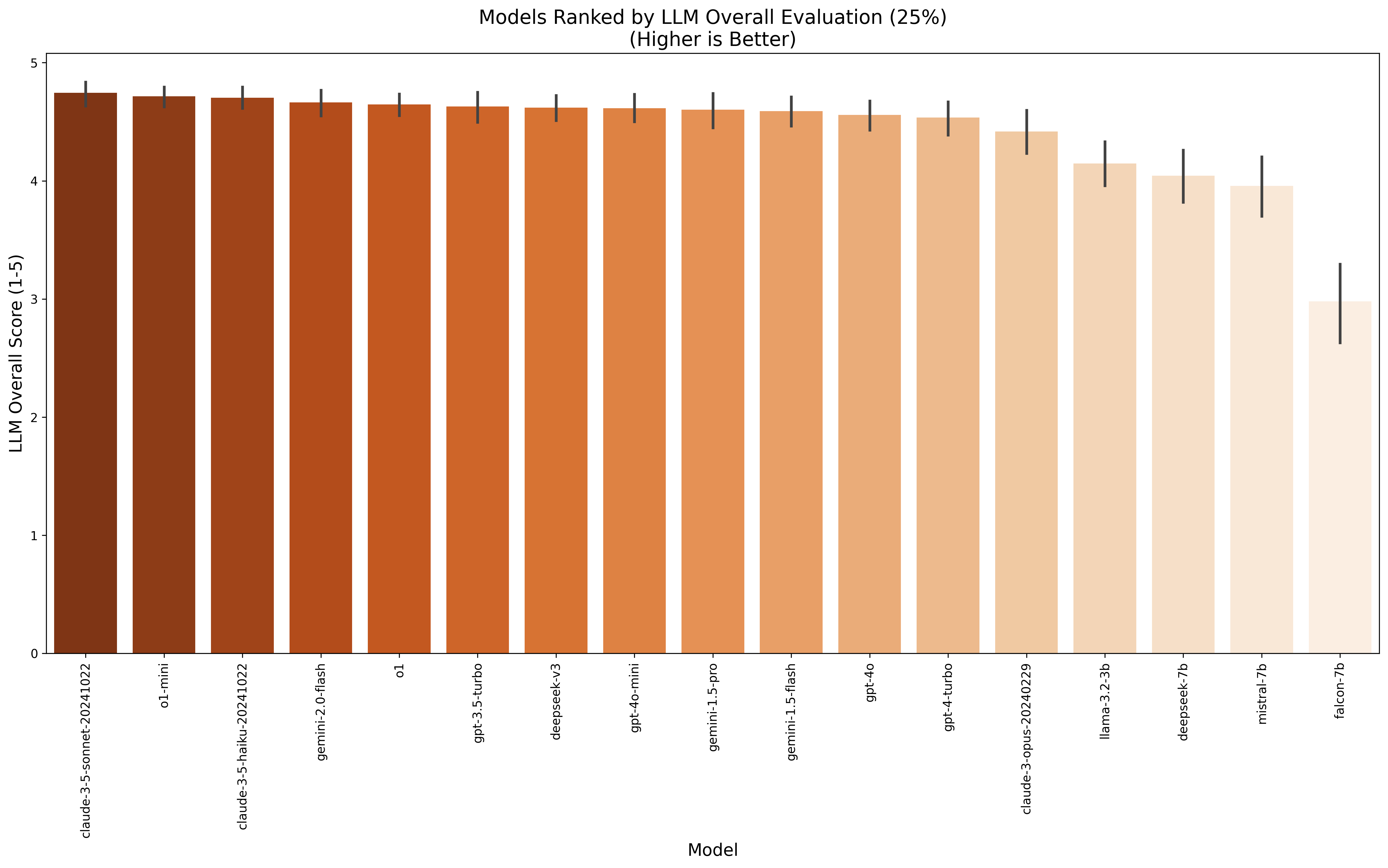}
    \caption{Models ranked by LLM evaluation scores}
    \label{fig:llm-eval-ranking}
\end{figure}

Figure \ref{fig:llm-eval-ranking} shows models ranked by human-like LLM evaluation scores, providing insight into perceived quality.

\section{Cost-Quality Tradeoff Multidimensional Visualization}

\begin{figure}[H]
    \centering
    \includegraphics[width=0.75\textwidth]{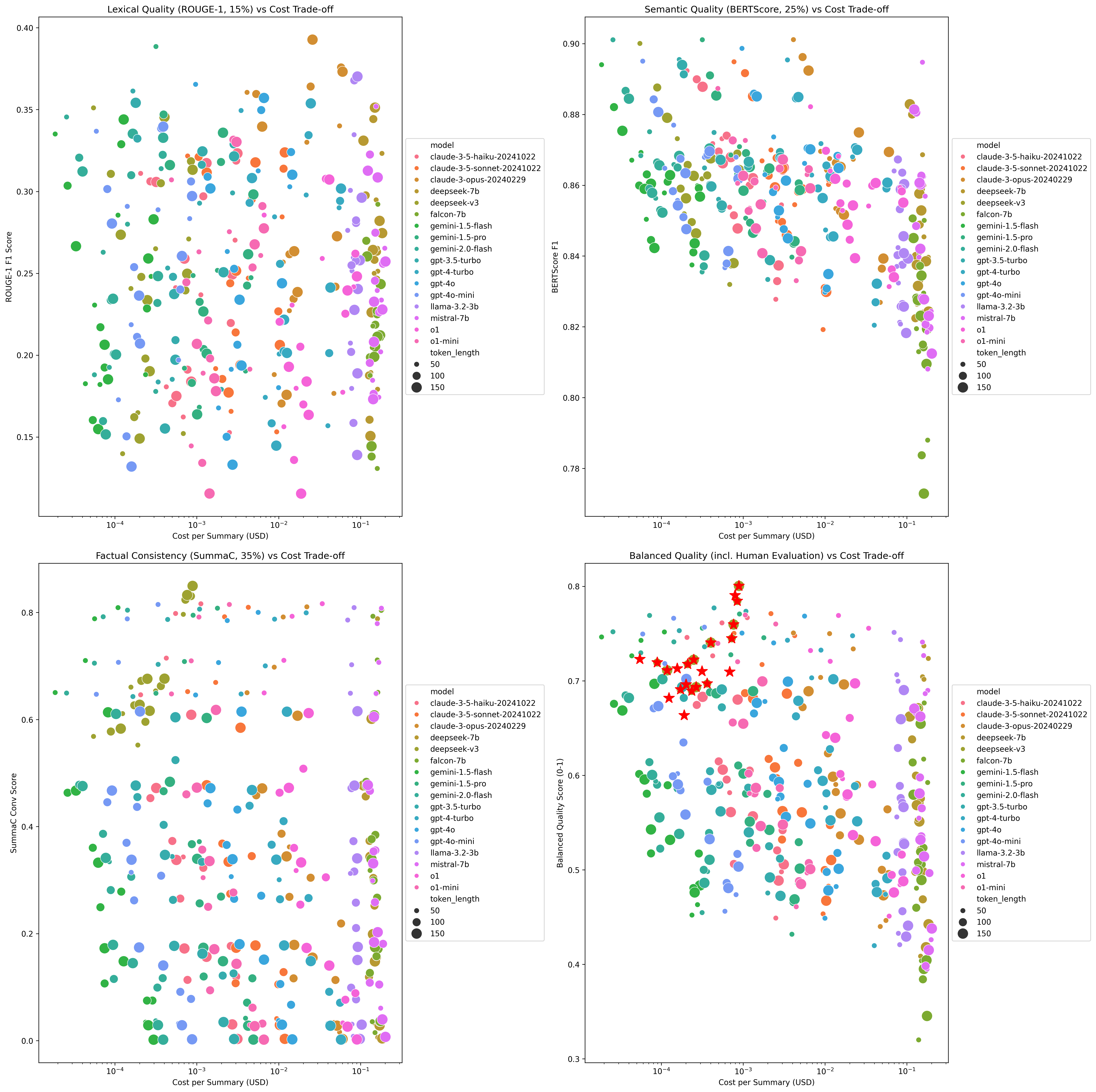}
    \caption{Multidimensional cost-quality trade-off across models}
    \label{fig:cost-quality}
\end{figure}

Figure \ref{fig:cost-quality} shows the multidimensional trade-off between quality aspects and cost across models, helping to identify the most cost-effective options for different budget constraints.

\section{Cost-Effectiveness Rankings}

\begin{figure}[H]
    \centering
    \includegraphics[width=0.75\textwidth]{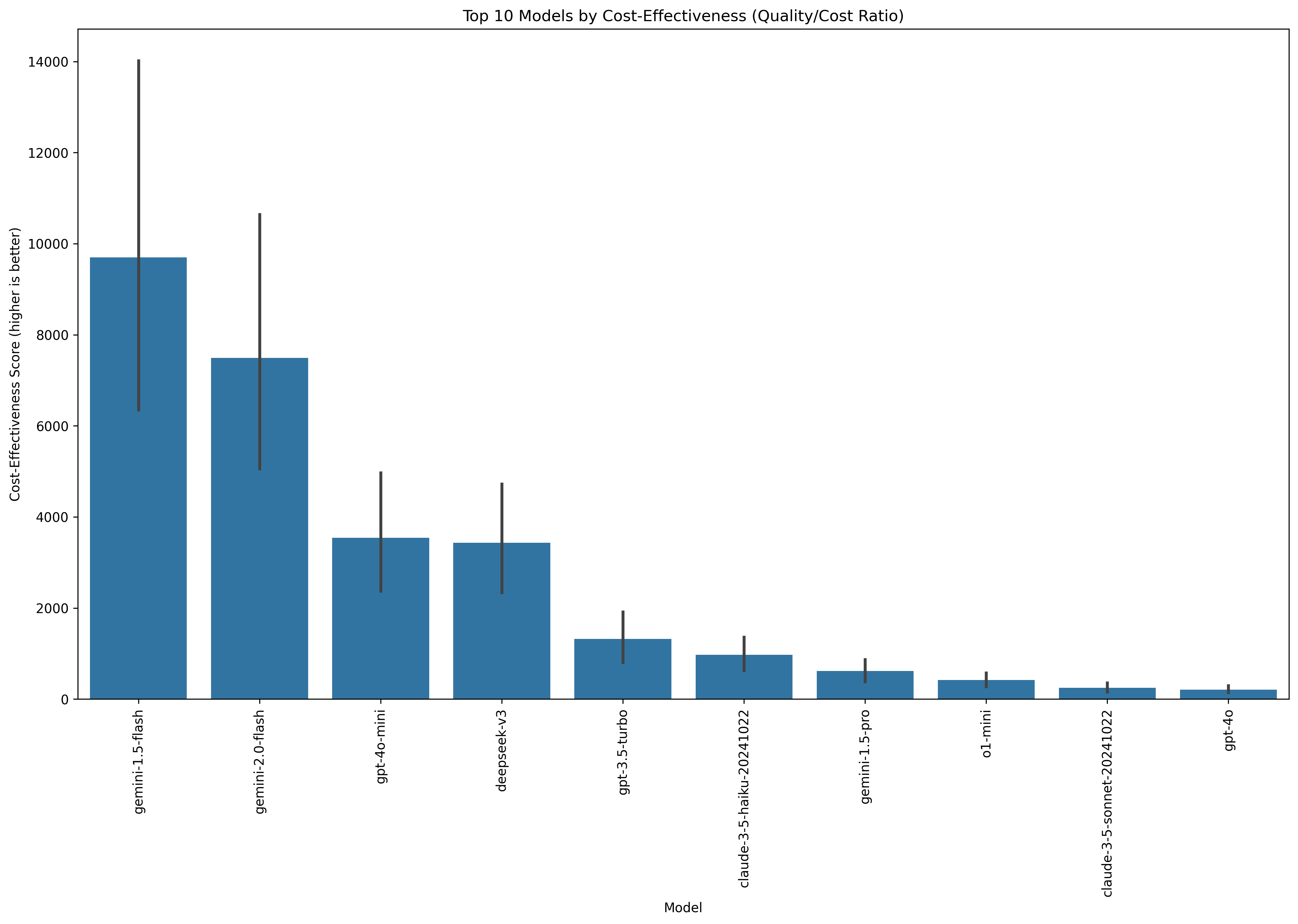}
    \caption{Models ranked by cost-effectiveness (quality per dollar)}
    \label{fig:cost-effectiveness}
\end{figure}

Figure \ref{fig:cost-effectiveness} presents models ranked by cost-effectiveness, identifying options that provide the best value for money.

\section{Dataset-Specific Optimal Model Configuration}

Table \ref{tab:dataset-optimal} identifies the best-performing model and token length configuration for each dataset, providing guidance for domain-specific applications.

\begin{table}[H]
\centering
\small
\begin{tabular}{@{}lcc@{}}
\toprule
\textbf{Dataset} & \textbf{Best Model} & \textbf{Optimal Token Length} \\
\midrule
CNN/DailyMail & gpt-3.5-turbo & 100 \\
XSum & deepseek-v3 & 50 \\
BigPatent & deepseek-v3 & 50 \\
BillSum & gpt-3.5-turbo & 100 \\
PubMed & deepseek-v3 & 50 \\
SAMSum & claude-3-5-haiku & 100 \\
WikiHow & gemini-2.0-flash & 100 \\
\bottomrule
\end{tabular}
\caption{Optimal model and token length configuration by dataset}
\label{tab:dataset-optimal}
\end{table}

\section{Quality-Efficiency Trade-off by Use Case}

Table \ref{tab:use-case} summarizes recommended model configurations for different use cases, balancing quality and efficiency requirements.

\begin{table}[H]
\centering
\small
\begin{tabular}{@{}lcc@{}}
\toprule
\textbf{Use Case} & \textbf{Recommended Model} & \textbf{Token Length} \\
\midrule
High-stakes (factual accuracy critical) & deepseek-v3 & 50 \\
Human-like quality & claude-3-5-sonnet & 150 \\
General-purpose & gpt-3.5-turbo & 100 \\
Resource-constrained & gemini-1.5-flash & 50-100 \\
Cost-effective & gemini-1.5-flash & 100 \\
\bottomrule
\end{tabular}
\caption{Recommended model configurations by use case}
\label{tab:use-case}
\end{table}

\section{Dataset-Specific Performance Visualizations}

The following figures illustrate the performance of models across different datasets, highlighting the significant performance variations by domain. These visualizations show how models perform differently on various text types and styles.

\subsection{CNN/DailyMail Dataset Performance}
\begin{figure}[H]
    \centering
    \includegraphics[width=0.90\textwidth]{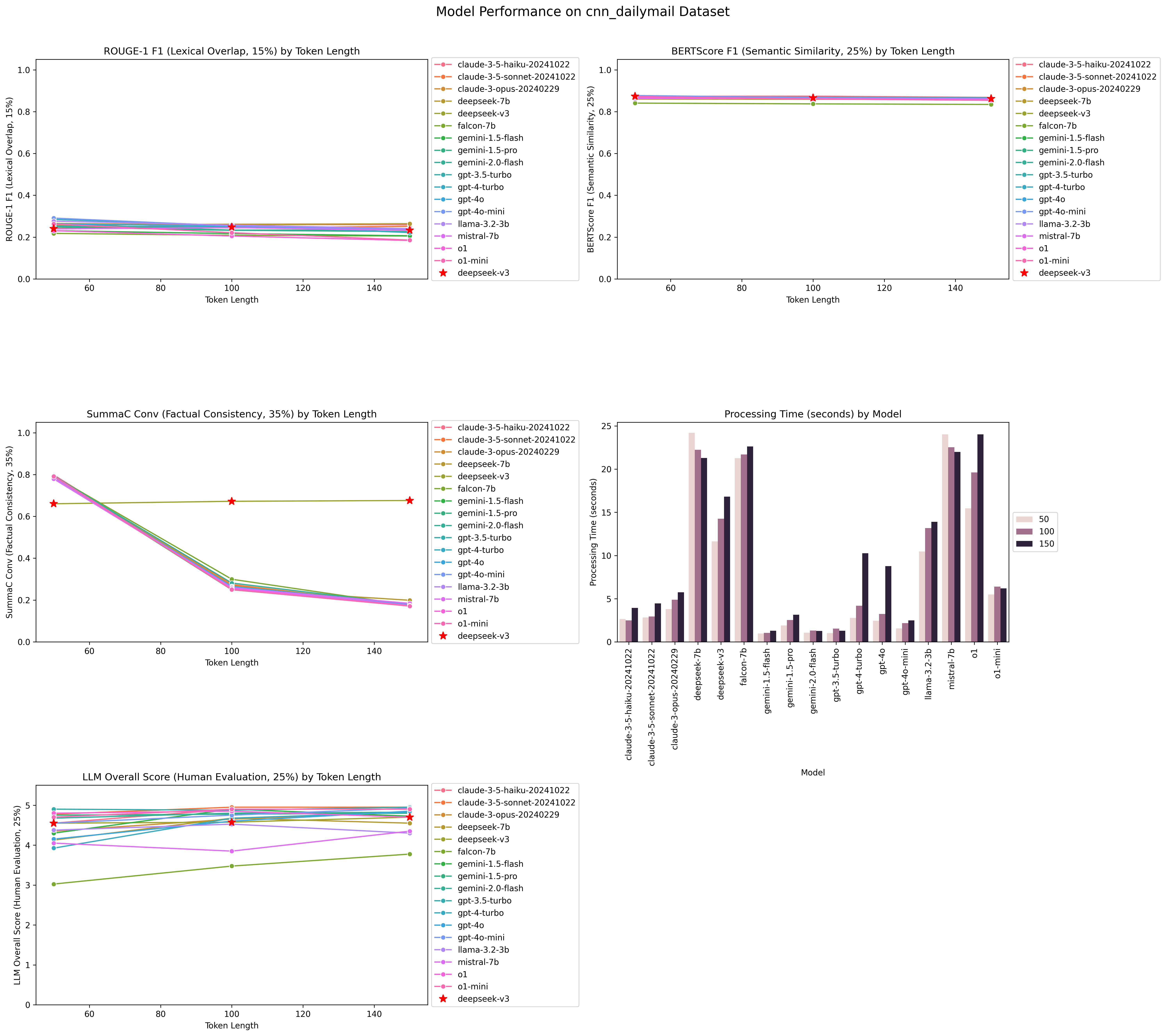}
    \caption{CNN/DailyMail dataset performance across metrics and models}
    \label{fig:cnn-performance}
\end{figure}

\subsection{XSum Dataset Performance}
\begin{figure}[H]
    \centering
    \includegraphics[width=0.95\textwidth]{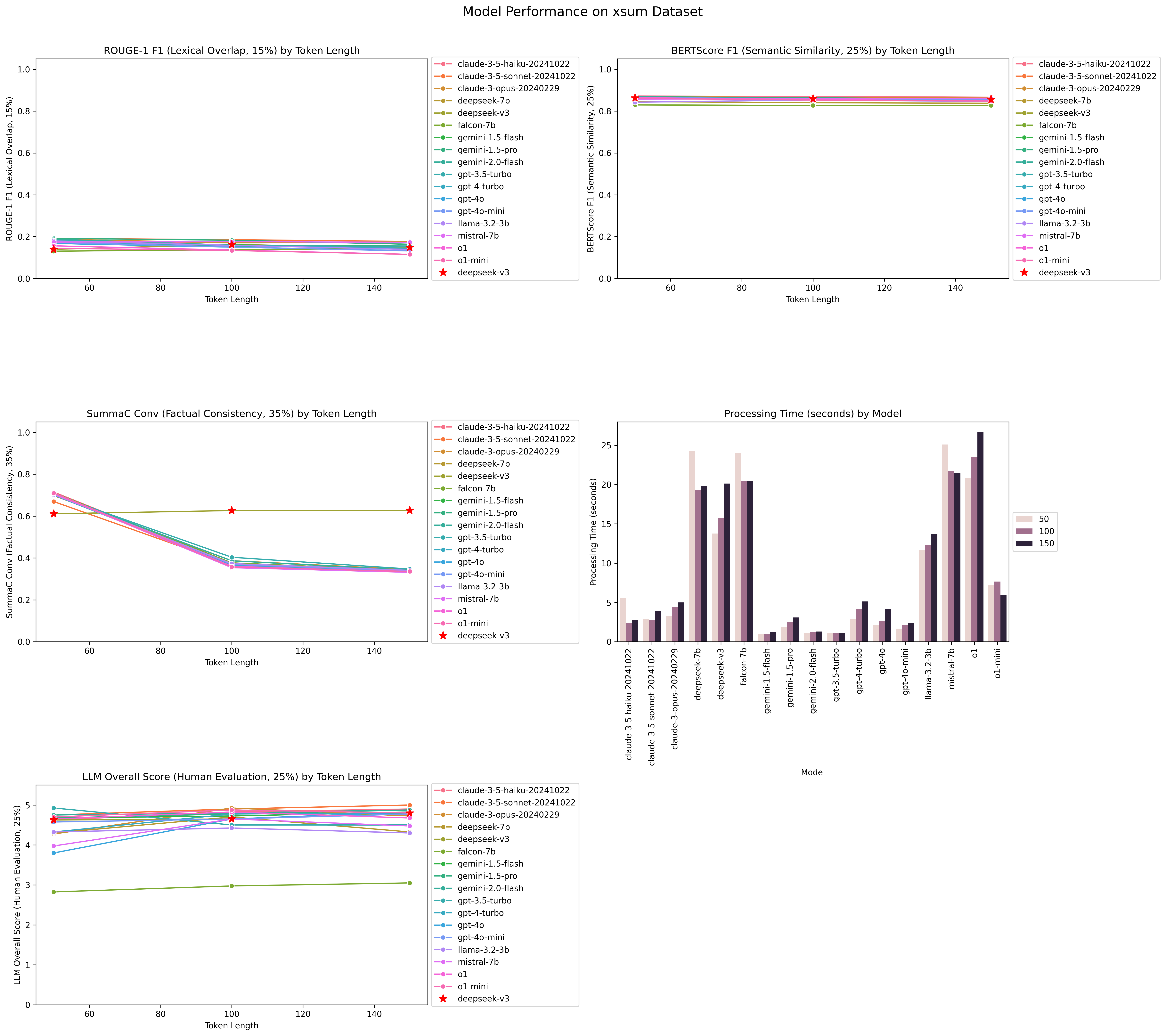}
    \caption{XSum dataset performance across metrics and models}
    \label{fig:xsum-performance}
\end{figure}

\subsection{SAMSum Dataset Performance}
\begin{figure}[H]
    \centering
    \includegraphics[width=0.95\textwidth]{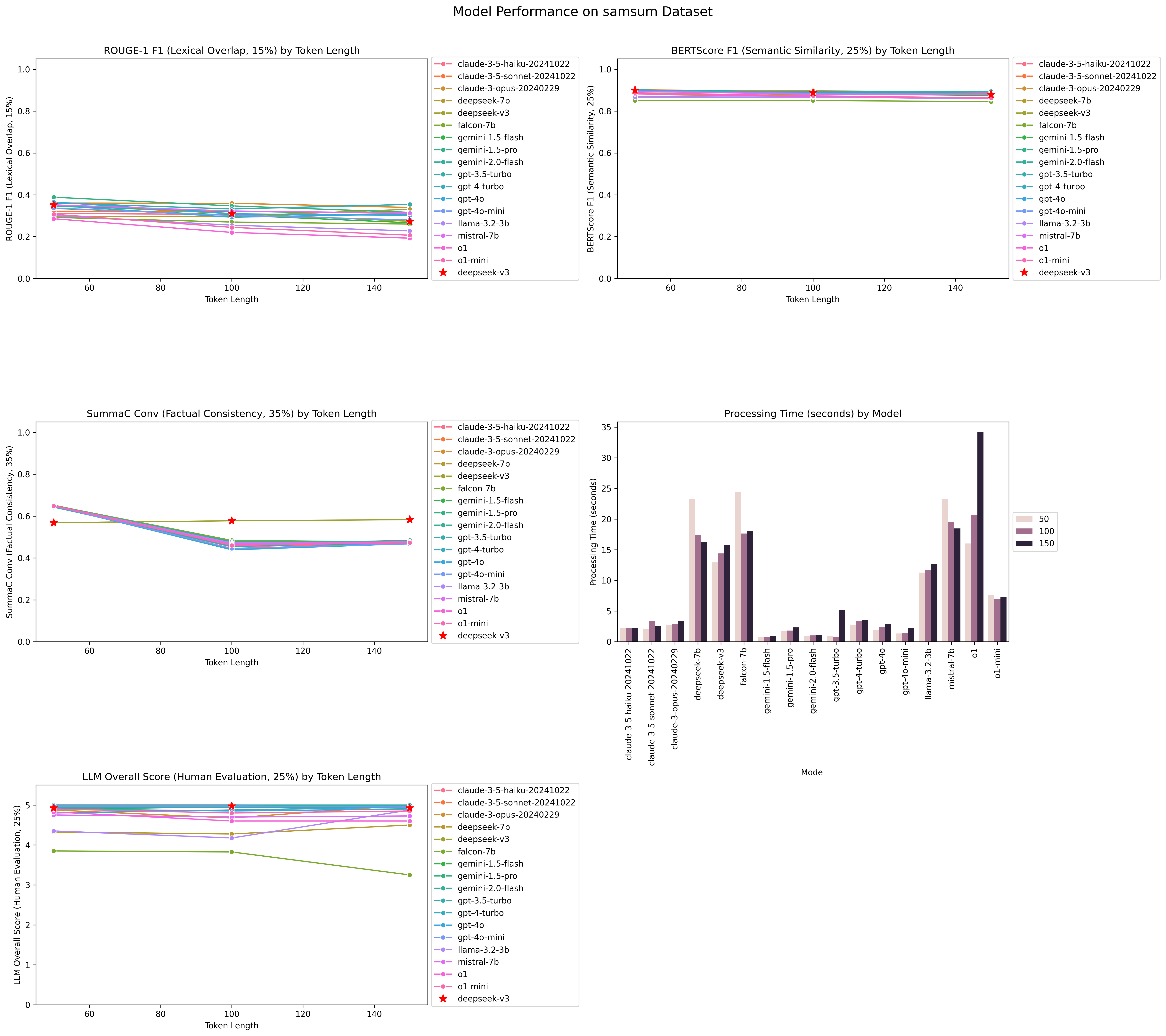}
    \caption{SAMSum dataset performance across metrics and models}
    \label{fig:samsum-performance}
\end{figure}

\subsection{PubMed Dataset Performance}
\begin{figure}[H]
    \centering
    \includegraphics[width=0.95\textwidth]{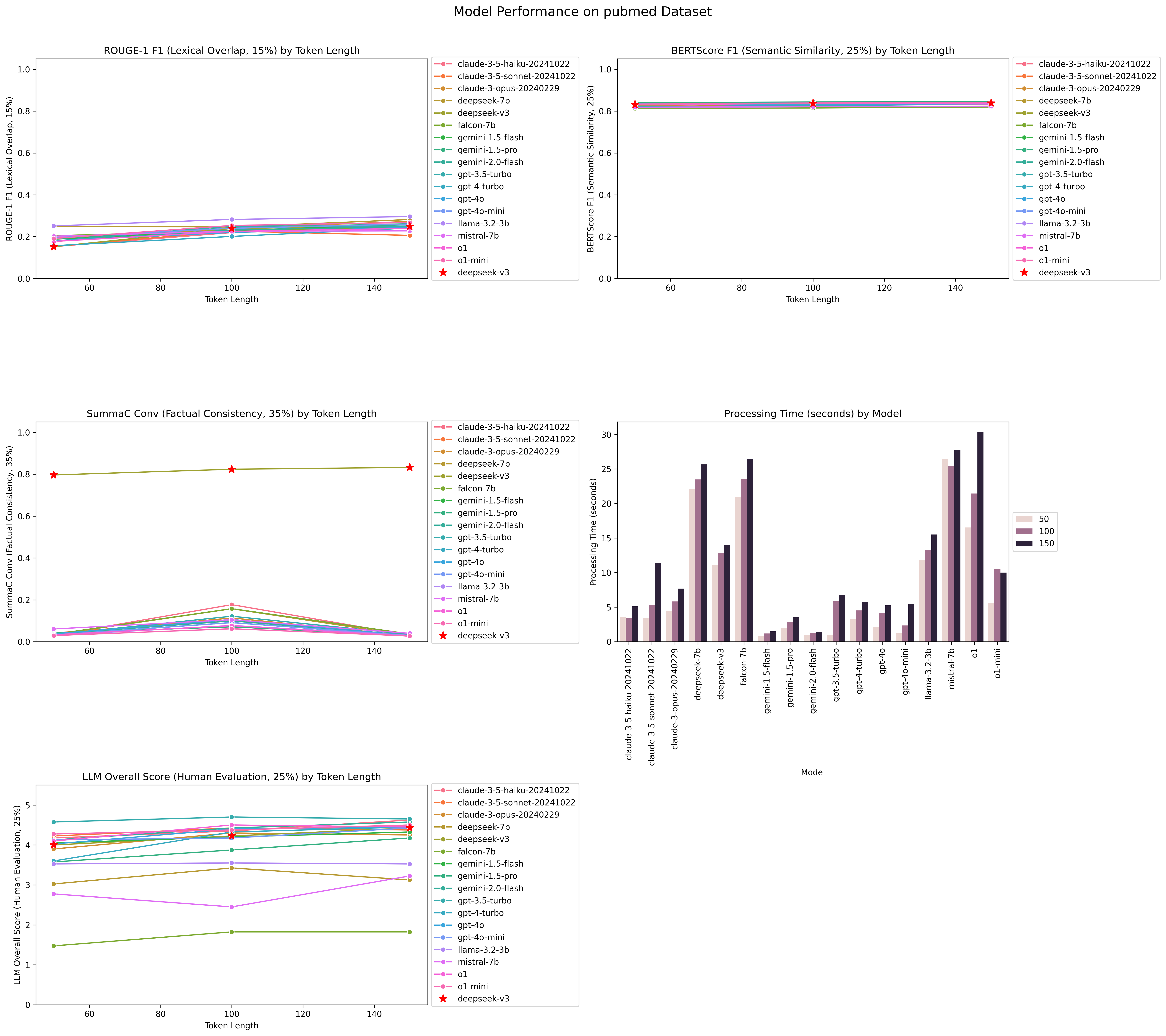}
    \caption{PubMed dataset performance across metrics and models}
    \label{fig:pubmed-performance}
\end{figure}

\subsection{BigPatent Dataset Performance}
\begin{figure}[H]
    \centering
    \includegraphics[width=0.95\textwidth]{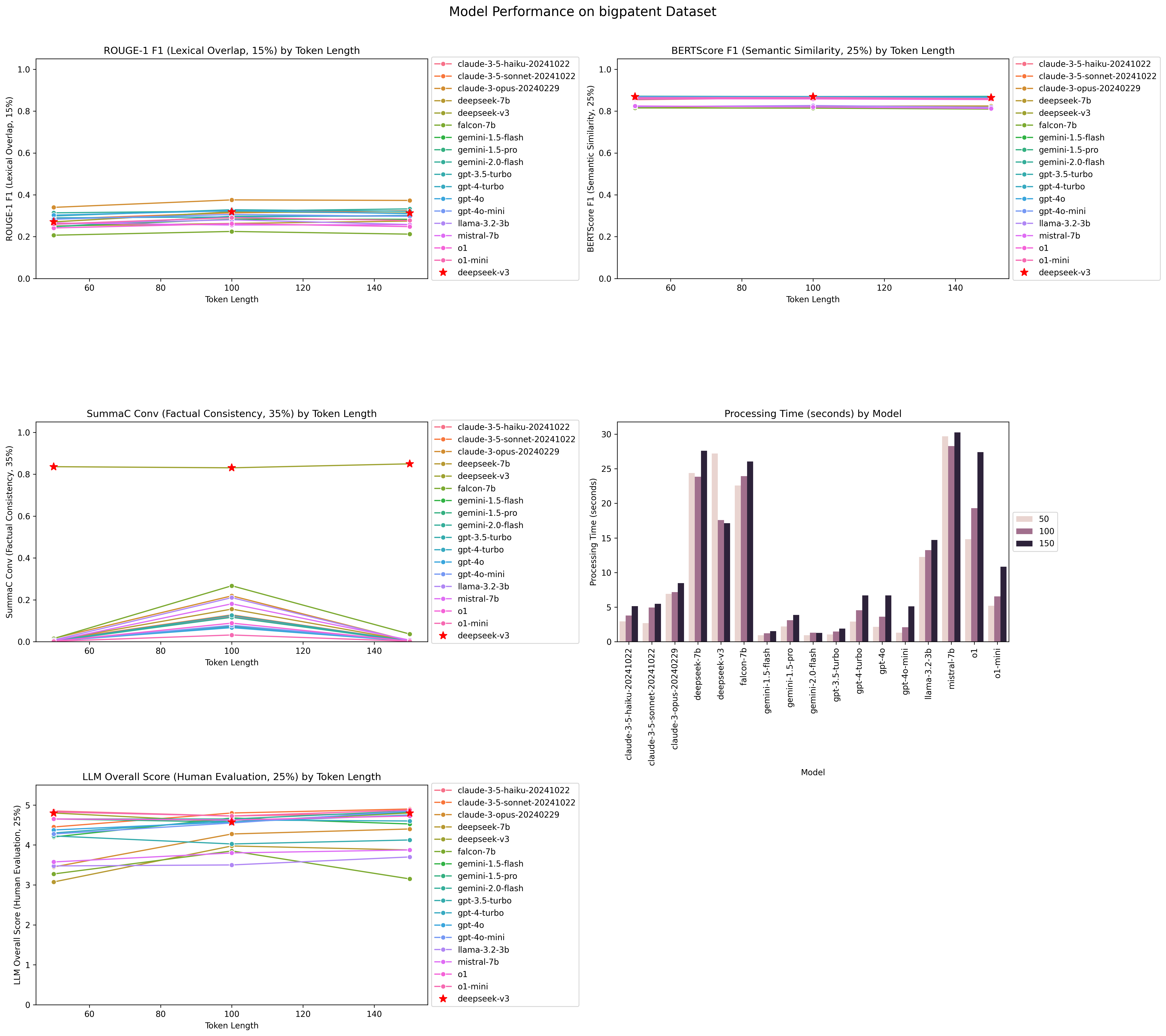}
    \caption{BigPatent dataset performance across metrics and models}
    \label{fig:bigpatent-performance}
\end{figure}

\subsection{BillSum Dataset Performance}
\begin{figure}[H]
    \centering
    \includegraphics[width=0.95\textwidth]{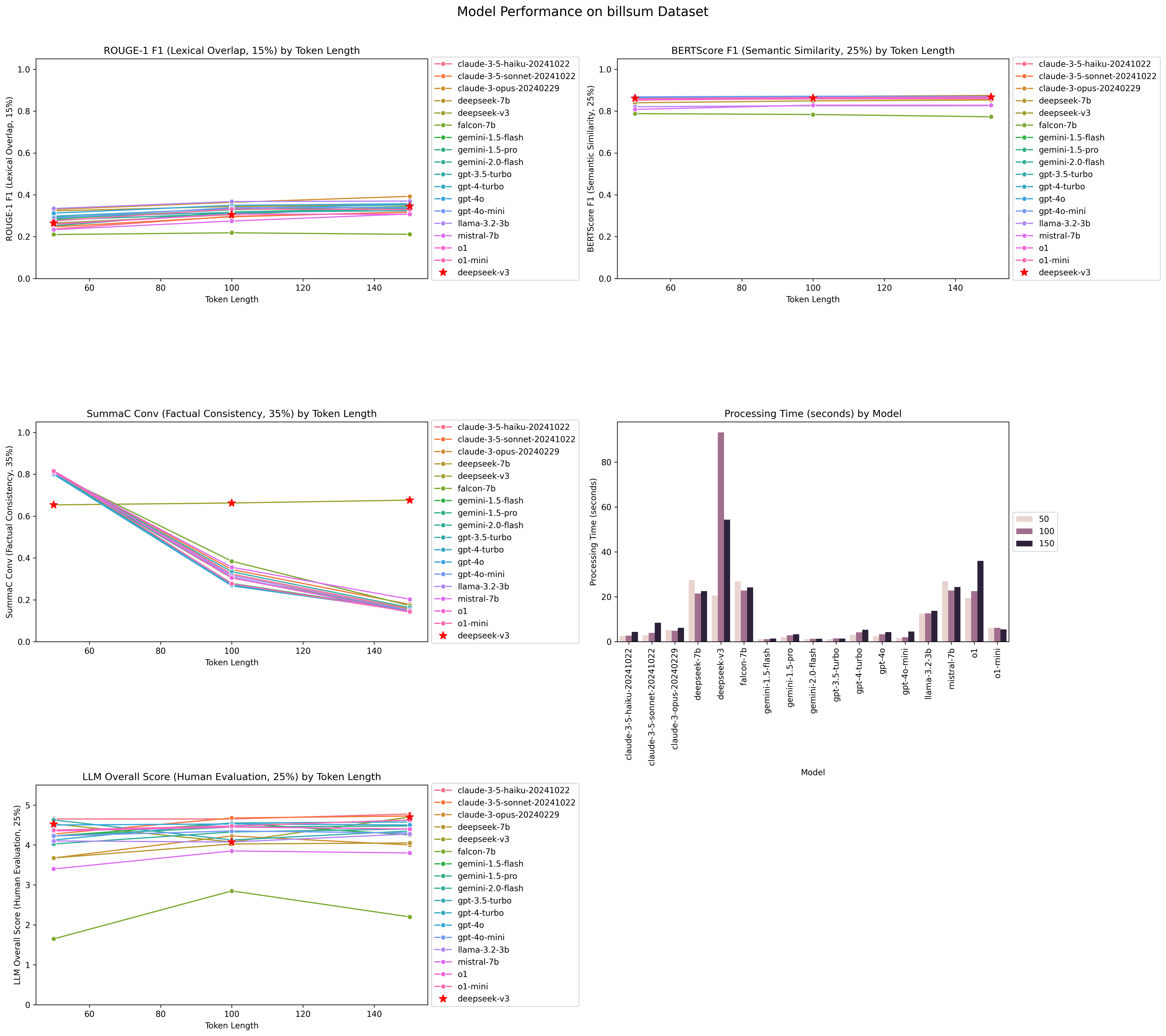}
    \caption{BillSum dataset performance across metrics and models}
    \label{fig:billsum-performance}
\end{figure}

\subsection{WikiHow Dataset Performance}
\begin{figure}[H]
    \centering
    \includegraphics[width=0.95\textwidth]{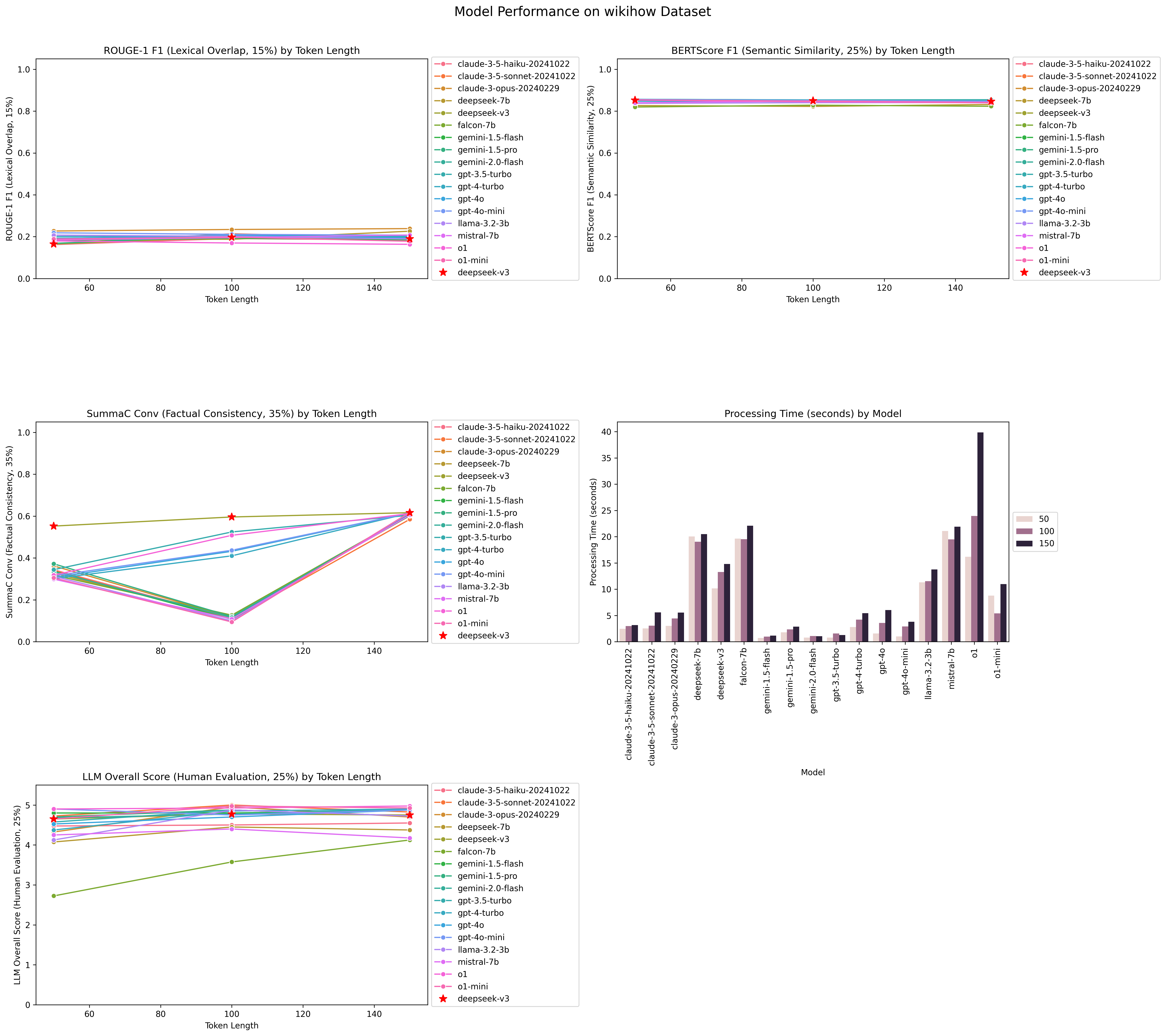}
    \caption{WikiHow dataset performance across metrics and models}
    \label{fig:wikihow-performance}
\end{figure}